\documentclass{article}

\PassOptionsToPackage{square,numbers, sort}{natbib} % 参考文献引用格式。参考：http://mirrors.ibiblio.org/CTAN/macros/latex/contrib/natbib/natbib.pdf

\usepackage[preprint]{neurips_2025}

\usepackage[utf8]{inputenc} % allow utf-8 input
\usepackage[T1]{fontenc}    % use 8-bit T1 fonts
\usepackage{hyperref}       % hyperlinks
\usepackage{url}            % simple URL typesetting
\usepackage{booktabs}       % professional-quality tables
\usepackage{amsmath}        % math formula
\usepackage{amsfonts}       % blackboard math symbols
\usepackage{nicefrac}       % compact symbols for 1/2, etc.
\usepackage{bm}             % bold math symbols
\usepackage{microtype}      % microtypography
\usepackage{graphicx}       % figure
\usepackage{subcaption}     % subfigure
\usepackage{multirow}       % table

\usepackage{algorithm}      % 算法浮动环境
\usepackage{algpseudocode}  % 伪代码语法
\usepackage{amsmath}

\usepackage{float}
\usepackage{amssymb}

\usepackage{caption}
\usepackage{bm}

\title{
A Brain-Inspired Gating Mechanism Unlocks Robust Computation in Spiking Neural Networks
}

\author{
  Baiqianyi~Bai\textsuperscript{1}\thanks{These authors contributed equally.}, \quad
  Haiteng~Wang\textsuperscript{2}\footnotemark[1], \quad
  Qiang~Yu\textsuperscript{1,3}\thanks{Corresponding author.} \\
  \textsuperscript{1}College of Intelligence and Computing, Tianjin University,Tianjin, China \\
  \textsuperscript{2}School of Future Technology, Tianjin University, Tianjin, China \\
  \textsuperscript{3}College of Computer and Information Engineering, Tianjin Normal University, Tianjin, China\\
  \texttt{\{baiqianyi, wanght\_0103, yuqiang\}@tju.edu.cn}
}

\begin{document}

\maketitle

\begin{abstract}
While spiking neural networks (SNNs) provide a biologically inspired and energy-efficient computational framework, their robustness and the dynamic advantages inherent to biological neurons remain significantly underutilized owing to oversimplified neuron models. In particular, conventional leaky integrate-and-fire (LIF) neurons often omit the dynamic conductance mechanisms inherent in biological neurons, thereby limiting their capacity to cope with noise and temporal variability. In this work, we revisit dynamic conductance from a functional perspective and uncover its intrinsic role as a biologically plausible gating mechanism that modulates information flow. Building on this insight, we introduce the Dynamic Gated Neuron~(DGN), a novel spiking unit in which membrane conductance evolves in response to neuronal activity, enabling selective input filtering and adaptive noise suppression. We provide a theoretical analysis showing that DGN possess enhanced stochastic stability compared to standard LIF models, with dynamic conductance intriguingly acting as a disturbance rejection mechanism. DGN-based SNNs demonstrate superior performance across extensive evaluations on anti-noise tasks and temporal-related benchmarks such as TIDIGITS and SHD, consistently exhibiting excellent robustness. Our results highlight, for the first time, a biologically plausible dynamic gating as a key mechanism for robust spike-based computation, providing not only theoretical guarantees but also strong empirical validations. This work thus paves the way for more resilient, efficient, and biologically inspired spiking neural networks.
\end{abstract}

\section{Introduction}
% 介绍SNN背景，并引出传统的SNN都是Gateless的
% 脉冲神经网络 (SNN) 利用离散的、事件驱动的脉冲进行计算，为传统神经架构提供了一种受生物学启发的替代方案。其能量效率、时间表达能力和抗噪性使其在神经形态应用中越来越有吸引力 \citep{maass1997networks, pfeiffer2018deep, roy2019towards, he2020comparing, cheng2020lisnn}。作为第三代网络，SNN 旨在弥合人工智能计算与生物现实之间的差距 \cite{pei2019towards}。然而，尽管具有生物学动机，大多数现有的 SNN 模型（通常称为无门 SNN）缺乏用于调节神经元动力学的内部门控机制。最近的研究，例如门控 LIF (GLIF) 模型，引入了静态的、通道级的门控，但在生物学上仍然难以实现。因此，该领域仍然缺乏针对脉冲神经元的生物学基础动态门控机制——这是开发更具适应性和鲁棒性的 SNN 的根本障碍。
Spiking Neural Networks (SNNs) offer a biologically inspired alternative to traditional neural architectures by leveraging discrete, event-driven spikes for computation. Their energy efficiency, temporal expressiveness, and robustness to noise make them increasingly attractive for neuromorphic applications \citep{maass1997networks, pfeiffer2018deep, roy2019towards, he2020comparing, cheng2020lisnn}. As third-generation networks, SNNs aim to bridge the gap between artificial computation and biological realism \cite{pei2019towards}. However, despite their biological motivations, most existing SNN models—commonly referred to as Gateless SNNs—lack internal gating mechanisms for modulating neuronal dynamics\cite{bellec2018alif, fang2021plif, zhang2024tclif}. Recent efforts, such as the Gated LIF (GLIF) model \cite{yao2022glif}, introduce static, channel-wise gates but remain biologically implausible. As a result, the field still lacks a biologically grounded dynamic gating mechanism for spiking neurons—a fundamental obstacle to developing more adaptive and robust SNNs.

% 关于动态电导的生物学基础
% 早期研究发现，生物神经元中的离子通道电导并不是静态的，而是可以通过依赖神经活动的生化过程在较慢的时间尺度上进行调节。研究表明，蛋白质磷酸化和基因表达可以调控离子通道的密度和功能，这种调节通常是在持续神经活动的背景下发生的 \cite{kaczmarek1987role, chad1986enzymatic, morgan1991stimulus, sheng1990regulation}。例如，去极化可诱导即时早期基因（如 fos 和 ras）的表达，从而引发钾通道电导的变化 \cite{smeyne1992fos}，而长时间的去极化则被观察到会降低钙电流 \cite{sharp1993dynamic}。这些过程通常由细胞内钙离子介导，它作为第二信使，将神经活动与通道电导的变化联系起来 \cite{kaczmarek1987neuromodulation}。这些生物发现激发了在计算模型中引入膜电导动态调节机制的研究，其中最具代表性的是Abbott和LeMasson提出的基于钙依赖机制的模型 \cite{abbott1993analysis}。除了在维持稳态中的作用之外，这种电导调节还可以被视为一种内在的门控机制，其通过膜特性对输入响应进行动态调控，并依据神经元的活动历史进行调整。这种具有生物学基础的门控方式独立于突触传递，在神经计算的调节中发挥着核心作用。 
% Early studies revealed that ion channel conductance in biological neurons are not static but can be modulated over slow timescales via activity-dependent biochemical processes. Protein phosphorylation and gene expression have been shown to regulate both the density and functionality of ion channels, typically in response to sustained neural activity \cite{kaczmarek1987role, chad1986enzymatic, morgan1991stimulus, sheng1990regulation}.
Protein phosphorylation and gene expression have demonstrated that ion channel conductance in biological neurons are not static; they can be dynamically modulated in response to sustained neural activity \cite{kaczmarek1987role, chad1986enzymatic, morgan1991stimulus, sheng1990regulation}. For instance, depolarization can trigger the expression of immediate early genes such as fos and ras, leading to changes in potassium conductance \cite{smeyne1992fos}, while prolonged depolarization has been observed to reduce calcium currents \cite{sharp1993dynamic}. Intracellular calcium often mediates these processes, serving as a second messenger that links neuronal activity to conductance modulation \cite{kaczmarek1987neuromodulation}. These findings have inspired computational models that incorporate dynamic regulation of membrane conductance, most notably the calcium-dependent framework proposed by Abbott and LeMasson \cite{abbott1993analysis}. Beyond their role in homeostatic adaptation, such conductance modulations can be viewed as intrinsic gating mechanisms, wherein membrane properties dynamically shape neuronal responsiveness based on prior activity. This biologically grounded form of gating operates independently of synaptic transmission and plays a central role in regulating neural computation.

% 在我们的模型中，膜电导会根据输入动态进行调节，使神经元能够控制跨膜信息流的幅度和时序。这一过程实现了一种生物合理的门控机制，可根据输入动态自适应地调控内部状态的持续性，使脉冲神经元能够控制历史信息的保留与衰减速度——在功能上类似于 LSTM 中的遗忘门。该过程实现了一种生物学上合理的门控机制，可根据输入动态自适应地调节内部状态的保持与衰减，使脉冲神经元能够控制过去信息的保留与遗忘，其功能上类似于LSTM中的遗忘门等循环结构中的门控操作 \cite{hochreiter1997long}。尽管LSTM和GRU等模型 \cite{cho2014learning} 通过人工设计的门控机制取得了优异表现，但它们的结构在很大程度上与生物机制脱节。我们的方法通过基于神经生理学原理构建门控动态，有效弥合了这一差距，并提供了将脉冲神经模型与人工门控循环单元联系起来的统一理论框架。这一生物启发的视角不仅提升了人工门控机制的可解释性，也促进了更具鲁棒性和适应性的架构设计，其灵感源于真实神经系统的动态特性。
Building on a series of biologically grounded studies on conductance-based neurons—including the influential work by Gütig \cite{gutig2009time}, which formulated their dynamic equations and revealed their time-warp–invariant property—we revisit this class of models to bridge the gap between biologically inspired dynamics and their underexplored integration into spiking neural network frameworks. We reintroduced the dynamic conductance mechanism into the LIF neuron model and proposed the Dynamic Gated Neuron (DGN) model. In DGN model, membrane conductance are dynamically modulated as a function of incoming activity, enabling the neuron to regulate the magnitude and timing of information flow across its membrane. This process implements a biologically plausible gating mechanism that adaptively modulates the persistence of internal states based on input dynamics, allowing spiking neurons to control the retention and decay of past information—functionally analogous to gating operations in recurrent architectures such as the forget gate in LSTMs \cite{hochreiter1997long}. While models like LSTMs and GRUs \cite{cho2014learning} have achieved remarkable performance through engineered gating schemes, their designs are largely disconnected from biological mechanisms. By grounding gating dynamics in neurophysiological principles, our approach bridges this gap, offering a unifying theoretical framework that links spiking neural models with artificial gated recurrent units. This biologically inspired perspective not only enhances the interpretability of gating functions in artificial systems, but also promotes the development of more robust and adaptive architectures informed by the dynamics of real neural circuits.

% DGN 模型在模拟生物神经元中观测到的膜电导动态方面实现了重要突破。不同于传统 LIF 神经元采用固定衰减率与静态电导的简化方法，我们的模型引入了输入依赖的电导调控机制，使神经元能够有选择地保留关键信息、抑制无关或噪声输入，从而构建出一种生物合理的门控功能。我们在多层脉冲神经网络中评估了所提出的模型，模型在实现强大的分类性能的同时，还表现出更强的抗噪声和干扰能力。我们的贡献总结如下：
The DGN model represents a substantial advancement in simulating the biologically observed dynamics of membrane conductance. Unlike traditional LIF neurons, which simplify neural dynamics by using fixed decay rates and static conductance parameters, our model introduces input-dependent modulation of membrane conductance. This enables neurons to selectively retain relevant information while suppressing irrelevant or noisy inputs, thereby implementing a biologically plausible gating mechanism. We evaluate the proposed model within multi-layer spiking neural networks and it achieves strong classification performance while demonstrating stronger resistance to noise and perturbation. Our contributions are summarized as follows:
\begin{itemize}
    \item \textbf{DGN:} We propose the Dynamic Gated Neuron (DGN) model, a generalized spiking neuron model with a fully derived membrane potential formulation. Central to DGN is a dynamic conductance mechanism that functions as a biologically plausible gating mechanism, enabling adaptive control over information flow and memory retention within the neuron.
    % 我们分析了动态电导与LSTM等ANN网络中门控结构的相似性，提供了LSTM优秀性能的生物学理论解析视角。
    \item \textbf{Bridging Biologically Inspired Dynamics and Artificial Gating Mechanisms: }We identify functional parallels between dynamic conductance modulation in our model and gating mechanisms in LSTM networks, offering a biologically grounded perspective that helps bridge the gap between brain-inspired computation and artificial neural networks.
    \item \textbf{Robustness Analysis and Accuracy Results:} We present a complete theoretical analysis of the anti-perturbation properties arising from dynamic conductance mechanisms. In addition, we conduct anti-noise experiments on benchmark datasets using the DGN model, which consistently demonstrates strong performance across both audio and neuromorphic tasks. Notably, our model achieves state-of-the-art top-1 accuracy of 99.10\% on the TIDIGITS dataset.
\end{itemize}

\section{Related Work}

\subsection{Biological and Computational Parametric Neuron Models}
% Hodgkin-Huxley (HH) 模型 \cite{hodgkin1952quantitative} 提出了基于电导的高保真神经元动力学建模方式，通过电压门控离子通道精确模拟了动作电位的产生。然而，由于其计算开销较高，该模型在脉冲神经网络（SNN）中的应用受到限制，后者通常采用如 LIF \cite{koch1998methods} 和 SRM \cite{gerstner2014neuronal} 等简化模型。Izhikevich 的对比分析 \cite{izhikevich2004model} 强调了不同模型在生物拟真性与计算效率之间的权衡。尽管电导模型具备更丰富的动态特性，它们在 SNN 中的整合仍较为稀少。相比之下，我们的工作不仅继承了动态电导的生物基础，还将其作为功能性门控机制引入，为生物建模与现代计算模型之间建立了连接。
The Hodgkin-Huxley (HH) model \cite{hodgkin1952quantitative} introduced a biophysically detailed, conductance-based description of neuronal dynamics, capturing action potential generation via voltage-gated ion channels. Despite its accuracy, the computational cost of solving HH equations limited its adoption in spiking neural networks (SNNs), which typically rely on oversimplified models such as LIF \cite{koch1998methods} and SRM \cite{gerstner2014neuronal}. Izhikevich's comparative analysis \cite{izhikevich2004model} further highlighted the trade-offs between biological realism and efficiency across neuron models. While conductance-based models offer richer dynamic properties, their integration into SNNs remains rare. In contrast, our work leverages dynamic conductance not merely for biophysical fidelity, but as a functional gating mechanism, bridging biophysical modeling with modern computational frameworks.
To enhance the temporal modeling capacity of spiking neurons, recent studies have extended the classical LIF framework by incorporating more flexible parameterizations or biologically inspired mechanisms, leading to the emergence of Computational Parametric Spiking Neurons \cite{yao2022glif}. Representative models include the Adaptive LIF (ALIF) neuron \cite{bellec2018alif}, which introduces activity-dependent threshold adaptation; GLIF \cite{yao2022glif}, which embeds gating mechanisms to modulate membrane potential dynamics; Heterogeneous LIF \cite{perez2021neural}, which enables learnable membrane time constants; and FS-neuron \cite{stockl2021optimized}, which treats all membrane-related parameters as trainable, a recent model introduces a double-threshold mechanism to enable both positive and negative spike generation \cite{zhou2024rethinking}. While these models improve expressiveness through structural extensions or trainability, our approach is more biologically grounded: it incorporates dynamic conductance as a functional gating mechanism, enabling adaptive regulation of information flow and memory retention with competitive performance across tasks.

\subsection{Robustness on SNNs}    % 抗噪声
% 为了增强SNN对噪声和对抗性扰动的鲁棒性，先前的研究大致可分为三类：结构化建模、基于训练的策略和受生物启发的机制。结构化方法侧重于神经元级特性；例如，调整发放阈值和时间窗口会显著影响对抗性鲁棒性 \cite{el2021securing}，而精确的脉冲时序已被证明可以稳定时间表征 \cite{ding2023spike}。其他研究探讨了膜电位泄漏在LIF神经元中的作用，表明适当的调节可以抑制高频扰动 \cite{sharmin2020inherent, chowdhury2021towards}；最近的方法进一步提出了可学习且异构的泄漏因子，以自适应地调节跨时间步长的信息保留。基于训练的方法通过在学习过程中注入对抗样本 \cite{kundu2021hire} 或应用 Lipschitz 正则化来限制梯度敏感度 \cite{ding2022snn} 来提高鲁棒性，尽管它们通常依赖于静态输入编码并忽略时间动态。受生物启发的策略模仿了在神经系统中观察到的机制，例如引入随机门控来模拟生物随机性 \cite{ding2024enhancing}，或利用基于频率的编码通过在不同时间步骤过滤高频成分来模拟选择性注意 \cite{xu2024feel}。总之，这些努力强调了将稳健的训练目标与符合生物学的时间调控相结合以提高 SNN 鲁棒性的重要性。基于这些见解，我们的 DGN 模型提出了一个新颖的鲁棒性增强框架，该框架与生物神经动力学更加契合，这就是我们的动态电导方案。
To enhance the robustness of SNNs against noise and adversarial perturbations, prior works can be broadly categorized into three types: structural modeling, training-based strategies, and biologically-inspired mechanisms. Structural approaches focus on neuron-level properties; for instance, adjusting firing thresholds and temporal windows significantly affects adversarial robustness \cite{el2021securing}, and precise spike timing has been shown to stabilize temporal representations \cite{ding2023spike}. Other works have investigated the role of membrane potential leakage in LIF neurons, demonstrating that proper tuning can suppress high-frequency perturbations \cite{sharmin2020inherent, chowdhury2021towards}. Recent methods further propose learnable and heterogeneous leak factors to adaptively regulate information retention across time steps\cite{fang2021plif, perez2021neural, ding2024robust}. Training-based methods improve robustness by injecting adversarial examples during learning \cite{kundu2021hire} or applying Lipschitz regularization to limit gradient sensitivity \cite{ding2022snn}, but they typically rely on static input encoding and overlook temporal dynamics. Biologically inspired strategies mimic mechanisms observed in neural systems, such as introducing stochastic gating to emulate biological randomness \cite{ding2024enhancing}, or leveraging frequency-based encoding to simulate selective attention by filtering high-frequency components at different time steps \cite{xu2024feel}. Together, these efforts highlight the importance of combining robust training objectives with biologically-aligned temporal regulation to improve SNN robustness. Building on these insights, our DGN model proposes a novel robustness-enhancing framework that more closely aligns with biological neural dynamics, which is our dynamic conductance scheme.

\section{Methodology}

\subsection{Dynamic Gated Neuron Model}

Extensive research on neuronal conductance mechanisms has established diverse conductance-based models \cite{fitzhugh1961impulses, morris1981voltage, hille2001ion, wilson1972excitatory, gutig2009time}, enhancing biological plausibility beyond traditional LIF frameworks. The neuronal dynamics of a basic conductance-based neuron can be described by the following formula:
\begin{gather}
    \frac{dV}{dt} = -g_lV + \sum_i^Ng_i(E_i-V) \label{1001} \\
    \frac{dg_{i}}{dt} = -\frac{1}{\tau_\text{s}}g_{i} + C_i\sum_{j}\delta(t-t_i^j) \label{1002}
\end{gather}
where $g_l$ is the leak conductance. $g_i$ represents the conductance of $i$-th synapse. $N$ signifies the number of presynaptic afferent. $t_i^j$ indicating the arrival time of the $j$-th presynaptic spike of the $i$-th afferent neuron before time $t$. $\tau_{\text{s}}$ is the synaptic time constant. $C_i$ represents learning weights of conductance. $E_i$ represents the equilibrium potential of the $i$-th synapse. $E_i$ has excitatory synaptic values and inhibitory synaptic values.

% 受神经生物学机制（神经元通道电导动态变化）的启发，我们提出了一个动态门控神经元 (DGN) 模型，该模型实现了基于生物学的门控结构，以协调神经元的生物物理学和计算效率。该框架引入动态电导构成基本门控机制，模拟生物神经元的自适应信号整合，同时保留关键的信息保留特性。该模型的数学公式通过两个相互作用的门控成分来捕捉膜电位动态：输入依赖的突触电导和固有泄漏电导。膜电位 V(t) 的时间演化遵循微分方程：

Further analysis of conductance-based neuron models revealed that membrane conductance($g_l+\sum g_iE_i$) exhibits activity-dependent plasticity modulated by presynaptic spiking patterns. This synaptic-driven mechanisms precisely regulate the decay rate of membrane potential, thereby modulating neuronal memory efficiency and temporal integration properties. Inspired by this, we present a Dynamic Gated Neuron (DGN) model that implements a biologically grounded gating structure to reconcile neuronal biophysics with computational efficiency. This framework introduces dynamic conductance as a fundamental gating mechanism, emulating biological neurons' adaptive signal integration while preserving critical information retention properties. The model's mathematical formulation controls membrane potential dynamics through two interacting gating components: input-dependent synaptic conductance and intrinsic leak conductance. The temporal evolution of membrane potential $V$ obeys the differential equation:
\begin{gather}
    \tau_{\text{s}}\frac{dD_i}{dt} = -D_i+ z_i^t \label{221}\\
    \frac{dV}{dt} = -(g_l + \sum_i^NC_iD_i)V + \sum_i^NW_iD_i \label{222}
\end{gather}
where $z_i^t$ represents the input spike of the $i$-th synapse in time $t$.  $D_i$ describes the exponentially decaying synaptic current to soma of the $i$-th synapse. $W_i$ represents learning weights of input current. Detailed derivations of these neuronal dynamics are provided in Appendix \ref{app:AADN_model}. 

\begin{figure}[t]
    \centering
    \includegraphics[width=1\linewidth]{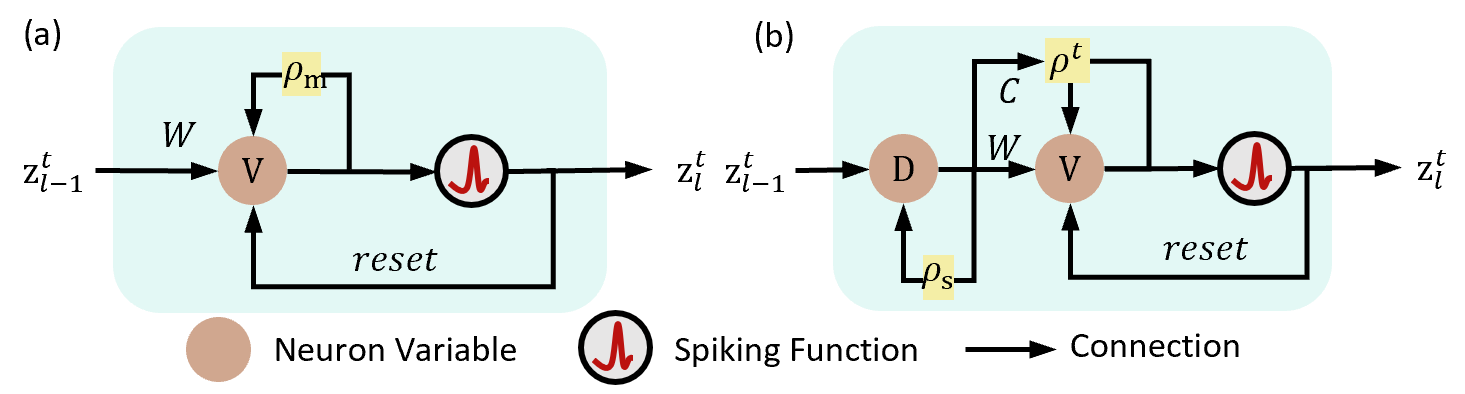}
    % \caption{Schematic of the neuron models (Eq.\eqref{223}-\eqref{226}). The standard leaky integrated-fire (LIF) model (a) uses a membrane decay factor \(\rho_m = e^{-\Delta t/\tau_m}\), which is controlled by the membrane time constant \(\tau_m\). The proposed DGN model (b) integrates synaptic inputs via \(\rho_s = e^{-\Delta t/\tau_s}\), where \(\tau_s\) represents the synaptic time constant. The total input current affects the membrane potential decay factor \(\rho^t\).}
    \caption{Schematic of the neuron models. (a) a standard LIF model(\(\rho_\text{m}=e^{-g_{l}\Delta t}\)). (b) the as-proposed DGN model described in Eq.\eqref{223}-\eqref{226}(\(\rho_\text{s}=e^{-\frac{\Delta t}{\tau_s}}\)).}
    \label{fig:neuron}
\end{figure}

For practical implementation of SNNs based on connected spiking neurons, coupled with spike firing and spike resetting processes, the dynamics of the DGN model are typically rendered in a discrete iterative format:
\begin{gather}
    D_i^t = e^{-\frac{\Delta t}{\tau_{\text{s}}}}D_i^{t-1} + z_i^t \label{223}\\[2mm]
    \rho^t = \varphi(1-g_l\cdot\Delta{t} -\Delta t\sum_i^NC_iD_i^t)\label{224}\\
    V^{t} = \rho^t\cdot V^{t-1} + \Delta t\sum_{i}^{N}W_iD_i^t - \vartheta z^{t-1} \label{225}\\
    z^t= \Theta(V^t-\vartheta) \label{226}
\end{gather}
 where $\Delta t$ represents the time interval between time steps in discrete form. $\varphi$ represents numerical truncation function, such as the Sigmoid function. $\Theta$ represents Heaviside step function. An output spike $z^t$ will be generated once the membrane potential $V^t$ reaches the neuronal firing threshold $\vartheta$ as per Eq.\eqref{226}. The membrane potential at the next time step will be soft reset as Eq.\eqref{225}.

This study compares the biologically inspired DGN with the LIF model to elucidate their structural distinctions (Fig. \ref{fig:neuron}(a-b)). Unlike the LIF model's fixed leakage conductance (\(g_l\)) and linear synaptic superposition, the DGN introduces dynamic conductance factors \(C_i\) to establish a dual-pathway regulatory architecture. It preserves the current injection pathway (\(W_iD_i\)) while adding a dynamic conductance term (\(C_iD_i\)), forming a gated mechanism governed by \(g_l+\sum C_iD_i\). This configuration can adaptively regulate the membrane potential decay rate in real time, thereby overcoming the limitations of LIF in simulating synaptic plasticity and increasing the efficiency of information transfer. This gated mechanism highlights DGN's advantages in balancing biophysical accuracy and computational performance. 

\subsection{Gating Structure Analysis in Conductance Dynamic Systems}

\begin{figure}
    \centering
    \includegraphics[width=1\linewidth]{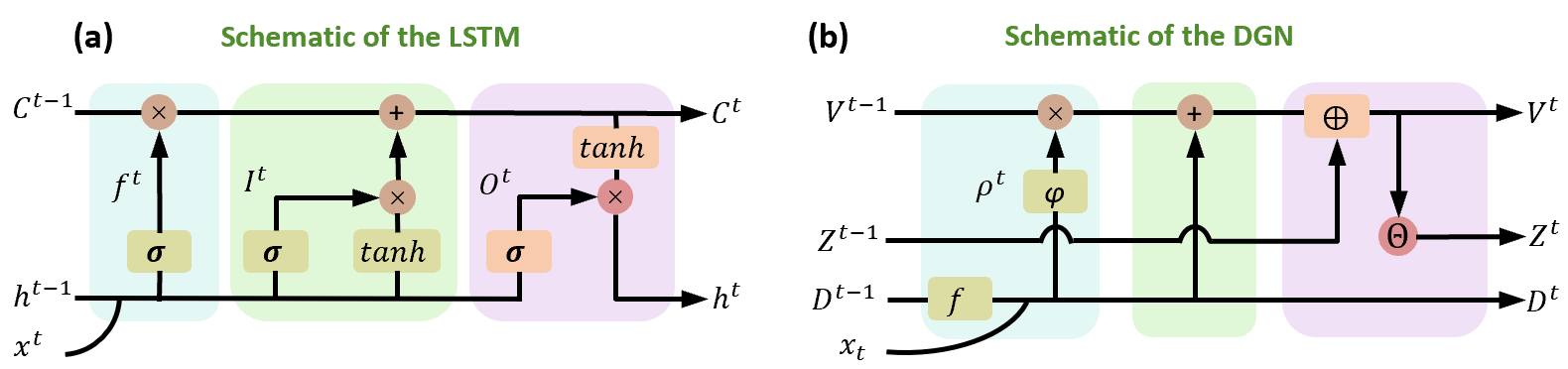}
    \caption{Schematic diagram of the model structure of LSTM and DGN. $f$: decay function. $\oplus$: reset processing.}
    \label{fig:rbla}
\end{figure}

% 所提出的动态门控神经元模型 (DGN) 在结构架构和信息处理机制上与长短期记忆 (LSTM) 网络表现出基本的拓扑同源性，如比较示意图所示。DGN 通过两项协同创新扩展了经典的 Leaky Integrate-and-Fire 框架：一个具有自适应系数的动态门控膜电位更新规则 (\(\rho^t\))，用于上下文敏感的历史记忆保留调节；以及一个突触加权机制，通过突触前整合动力学实现特征选择。这种生物学上可解释的架构与 LSTM 的三门机制具有显著的功能同构性——自适应衰减系数在数学上模拟了 LSTM 遗忘门 (\(f_t\)) 的记忆过滤功能，而突触加权机制则通过动态突触前整合在计算上与输入门控操作 (\(i_t\)) 相似。

% 如结构对比图所示，DGN 的脉冲重置机制与 LSTM 的细胞状态更新方程在数学上一致，两者都采用非线性门控变量来控制状态转换。这种拓扑一致性为 LSTM 的运行原理提供了神经生物学验证，表明其门控单元构成了神经元离子通道动力学的计算抽象，而非纯粹的工程解决方案。具体而言，生物门控通道的电压依赖性激活/失活特性在 LSTM 的 S 形门控函数中找到了对应的数学模型，这通过生物物理现实性解释了其卓越的时间处理能力。

The proposed Dynamic Gated Neural (DGN) model demonstrates fundamental topological homology with Long Short-Term Memory (LSTM) networks in both structural architecture and information processing mechanisms, as illustrated in Fig.~\ref{fig:rbla}. Our DGN employs an adaptive coefficient ($\rho^t$) to dynamically regulate membrane potential and memory retention while accumulating presynaptic currents for enhanced feature selection. This biologically interpretable architecture reveals striking functional isomorphism with LSTM's triple-gate mechanism - the self-adapting decay coefficient mathematically emulates the memory filtration function of LSTM's forget gate (\(f^t\))\cite{hochreiter1997long}. The mechanism of accumulating currents through dynamic presynaptic integration\cite{gerstner2014neuronal} is computationally similar to the input gating operation (\(I^t\)).

% \cite{bellec2020solution}
As shown in the Fig. \ref{fig:rbla}, DGN's spike reset mechanism exhibits mathematical congruence with LSTM's cell state update equations, both employing nonlinear gating variables to control state transitions. This topological alignment provides neurobiological validation for LSTM's operational principles, suggesting that its gating units constitute computational abstractions of neuronal ion channel dynamics\cite{hodgkin1952quantitative} rather than purely engineered solutions. Specifically, the voltage-dependent activation/inactivation characteristics\cite{burkitt2006review} of biological gating channels find their mathematical counterparts in LSTM's sigmoidal gate functions, explaining its superior temporal processing capabilities.
%through biophysical realism\cite{dayan2001theoretical}.

% DGN 相对于 LIF 模型的进步与 LSTM 相对于 Vanilla RNN 的改进之间的演化对应性揭示了一种跨尺度的设计范式。这两种架构都通过门控介导的状态调制来解决时空信息处理的核心挑战——记忆保存和动态特征选择。如示意图所示，这种结构同源性建立了一个统一的框架，其中神经计算原理涵盖了生物系统和人工系统。 DGN 的膜电位门控机制在不同组织尺度上映射了 LSTM 的细胞状态控制，表明无论是在生物神经元还是人工单元中，有效的时间处理都需要功能等效的解决方案。

The evolutionary correspondence between DGN's advancement over LIF models and LSTM's improvement upon Vanilla RNNs\cite{bengio1994learning} reveals a cross-scale design paradigm. Both architectures address the core challenges of spatiotemporal information processing - memory preservation and dynamic feature selection - through gating state modulation. As visualized in Fig. \ref{fig:rbla}, this structural homology establishes a unified framework where neural computation principles span biological and artificial systems. DGN's membrane potential gating mechanism mirrors LSTM's cell state control at different organizational scales, demonstrating that effective temporal information processing requires functionally equivalent solutions whether in biological neurons or artificial units.

% 这项跨模型验证凸显了门控机制在时间信息处理中的普遍意义，同时建立了一种新的仿生设计范式。揭示的同构性表明，对神经元亚细胞动力学（尤其是电压门控离子通道相互作用）的更深入探索，可以产生具有更高生物合理性和计算效率的新型神经网络架构。我们的研究结果将 LSTM 的成功定位为一项工程学突破，更是对进化神经计算原理的无意识重现。

% This cross-model validation highlights the universal significance of gating mechanisms in temporal information processing.% while establishing a new bio-inspired design paradigm. 
% The revealed isomorphism suggests that deeper exploration of neuronal subcellular dynamics, particularly voltage-gated ion channel interactions\cite{hille2001ion}, could yield novel neural network architectures with enhanced biological plausibility and computational efficiency\cite{indiveri2011neuromorphic}. Our findings position LSTM's success not merely as an engineering breakthrough but as an unconscious recapitulation of evolved neural computation principles.

\subsection{Stability of Conductance Dynamic Systems}  
This section establishes the theoretical framework for analyzing noise robustness in DGN model through stochastic differential equation (SDE) approaches. By linearizing the nonlinear conductance dynamics under small perturbation assumptions, we derive closed-form expressions for steady-state voltage variances in both DGN and classical LIF model. Comparative analysis of these variance solutions reveals the superior noise suppression capability of the DGN architecture.  

In order to compare fairly with other models, we directly analyze the case of adding perturbations to the presynaptic input current ($D_i$ in Eq.(\ref{221})). The investigation begins with stochastic input perturbations modeled as Gaussian white noise superposed on deterministic signals:  
\begin{gather}
    \hat{I}_i(t) = \mu_i + \sigma_i \xi(t), \quad \langle \xi(t)\xi(t') \rangle = \delta(t-t') \label{331}
\end{gather}
where \(\mu_i\) denotes deterministic input components and \(\sigma_i\) quantifies noise intensity. The perturbed dynamic conductance \(\hat{G}(t) = G_0 + \sum C_i \sigma_i \xi(t)\) induces voltage dynamics, where \(G_0=g_l + \sum C_i\mu_i\). The membrane potential control formula is: 
\begin{equation}
\frac{dV}{dt} = -\hat{G}(t)V + \sum W_i \hat{I}_i(t) = \underbrace{-G_0V + \sum W_i \mu_i}_{\text{Deterministic term}} + \underbrace{(- \sum C_i \sigma_i \xi(t)V + \sum W_i \sigma_i \xi(t))}_{\text{Perturbation term}}\label{333}
\end{equation}
Linear noise approximation (LNA) is applied by decomposing \(V = V_{\text{steady}} + \delta V(t)\) with \(|\delta V| \ll V_{\text{steady}}\), where $V_{\text{steady}}$ is the steady-state solution of the deterministic term. Performing Taylor expansion on nonlinear terms $C_i \sigma_i \xi(t) V$ and retaining only first-order contributions while discarding higher-order small terms (\(\delta V \cdot \xi(t)\)). The nonlinear perturbation term is linearized as:
\begin{gather}
    C_i \sigma_i \xi(t) V \approx C_i \sigma_i \xi(t) V_{\text{steady}} \label{335}
\end{gather}
After truncating higher-order terms, the original SDE reduces to a linear SDE:
\begin{equation}
\frac{dV}{dt} = -G_0V + \sum W_i \mu_i +\sum \sigma_i \left( W_i - C_i V_{\text{steady}} \right) \xi(t) \label{336}
\end{equation}
Using Itô calculus\cite{ito1944109} %and Green's function methods\cite{gardiner1985handbook}, %
the steady-state variance for DGN resolves to:  
\begin{equation}
\langle V^2 \rangle_{\text{DGN}} = \frac{\left[\sum_{i=1}^N \sigma_i \left( W_i - \frac{C_i \sum_{j=1}^N W_j \mu_j}{G_0} \right)\right]^2}{2G_0} \label{337}
\end{equation}
For classical LIF neurons with constant leak \(g_l\), the corresponding variance reduces to:  
\begin{equation}
\langle V^2 \rangle_{\text{LIF}} = \frac{(\sum_{i=1}^N W_i \sigma_i)^2}{2g_l} \label{338}
\end{equation}

The derivation process of the above formula is detailed in \ref{app:SDE}. Critical examination of Eq.(\ref{337}) versus Eq.(\ref{338}) demonstrates two synergistic noise suppression mechanisms in DGN. The denominator \(G_0\) implements input-dependent leakage scaling, where intensified inputs \(\mu_i\) amplify effective conductance to suppress voltage fluctuations. The numerator contains a compensatory term \(\frac{C_i \sum W_j \mu_j}{G_0}\) that introduces negative feedback proportional to synaptic weights \(W_i\) and coupling coefficients \(C_i\). When \(W_i\) and \(C_i\) are positively correlated, this feedback cancels synaptic noise propagation through \(W_i\), achieving partial noise rejection. In contrast, the LIF model's fixed leakage \(g_l\) and absence of compensatory terms result in static noise scaling that cannot adapt to input statistics.  

These analytical results quantitatively demonstrate that DGN neurons outperform LIF models in noise resilience through dynamic conductance modulation. The dual mechanism—adaptive leakage scaling and synaptic noise compensation—enables effective voltage stabilization during concurrent signal and noise processing. This theoretical framework provides fundamental insights into how conductance dynamics enhance neural computation robustness under stochastic perturbations.

\begin{table}[t]
\centering
\small
% \caption{Comparison of methods across datasets with and without recurrent connections (Rec), including clean accuracy. Methods with * are results reproduced by us. Highlighted rows (blue background) correspond to our proposed method, while bold entries indicate the best performance.}
\caption{Comparison of model performance on Ti46Alpha, TIDIGITS, SHD, and SSC datasets. Rec=N/Y represents feedforward networks and recurrent networks, respectively. * indicates results we reproduced using public code, while bold entries indicate the best performance.}
\vspace{1mm}
\begin{tabular}{clccc}
\toprule
\textbf{Datasets} & \textbf{Method} & \textbf{Rec} & \textbf{Hidden Layers} & \textbf{Accuracy(\%)} \\ \midrule
\multirow{7}{*}{\rotatebox{90}{Ti46Alpha}} 
 & LIF + HM2-BP\textsubscript{NeurIPS, 2018\cite{jin2018hybrid}} & N & 800-800 & 90.98 \\
 & \textbf{DGN(Ours)} & N & \textbf{100} & \textbf{95.69} \\ \cmidrule{2-5} 
 & RNN* & Y & 100 & 91.89 \\
 & LSTM* & Y & 100 & 96.05 \\
 & LIF + SrSc-SNNs-BIP\textsubscript{Neural Comput., 2021\cite{zhang2021skip}} & Y & 
 % SrSc-SNNs-BIP(400-400-400) 
 400-400-400
 & 95.90 \\
 & \textbf{LIF + SrSc-SNNs-BIP\textsubscript{Front. Neurosci., 2024\cite{zhang2024composing}}} & Y & \textbf{
 % SrSc-SNNs-BIP()
 800
 } & \textbf{96.44} \\
 & DGN(Ours) & Y & 100 & 96.31 \\ \midrule

\multirow{7}{*}{\rotatebox{90}{TIDIGITS}} 
 & LIF + BAE-MPDAL\textsubscript{Front. Neurosci., 2020\cite{pan2020efficient}} & N & 620-11 & 97.40 \\
 & LIF + Multilayer FE-Learn\textsubscript{TNNLS, 2023\cite{luo2022supervised}} & N & 100-100 & 98.10 \\
 & LIF + BPTE\textsubscript{IJCNN, 2023\cite{lin2023bipolar}} & N & 400-11 & 98.10 \\
 & \textbf{DGN(Ours)} & N & \textbf{100} & \textbf{98.59} \\ \cmidrule{2-5} 
  & RNN* & Y & 100 & 97.09 \\
 & LSTM* & Y & 100 & 97.88 \\
 & \textbf{DGN(Ours)} & Y & \textbf{100} & \textbf{99.10} \\ \midrule

\multirow{10}{*}{\rotatebox{90}{SHD}} 
 & LIF + data aug\textsubscript{TNNLS, 2020\cite{cramer2020heidelberg}} & N & 128 & 49.70 \\
 & TC-LIF\textsubscript{AAAI, 2024\cite{zhang2024tclif}} & N & 128-128 & 83.08 \\
 & \textbf{DGN(Ours)} & N & \textbf{128} & \textbf{85.18} \\ \cmidrule{2-5} 
  & RNN* & Y & 100 & 76.53 \\
 & LSTM\textsubscript{TNNLS, 2020\cite{cramer2020heidelberg}} & Y & 128 & \textbf{89.20} \\
 & LIF + data aug\textsubscript{TNNLS, 2020\cite{cramer2020heidelberg}} & Y & 1024 & 84.50 \\
 & Heterogeneous LIF\textsubscript{Nat. Commun., 2021\cite{perez2021neural}} & Y & 128 & 83.50 \\
 & ALIF\textsubscript{TNNLS, 2020\cite{yin2020effective}} & Y & 128-128 & 84.40 \\
 & \textbf{TC-LIF\textsubscript{AAAI, 2024\cite{zhang2024tclif}}} & Y & \textbf{128-128} & \textbf{88.91} \\
 & DGN(Ours) & Y & 128 & 87.78 \\ \midrule

\multirow{11}{*}{\rotatebox{90}{SSC}} 
 & LIF\textsubscript{TNNLS, 2020\cite{cramer2020heidelberg}} & N & 128-128 & 38.50 \\
 & TC-LIF\textsubscript{AAAI, 2024\cite{zhang2024tclif}} & N & 128-128 & 63.46 \\
 & \textbf{DGN(Ours)} & N & \textbf{128-128} & \textbf{67.54} \\ \cmidrule{2-5} 
  & RNN* & Y & 128-128 & 72.91 \\
 & LSTM\textsubscript{TNNLS, 2020\cite{cramer2020heidelberg}} & Y & 128-128 & 73.10 \\
  & LIF\textsubscript{TNNLS, 2020\cite{cramer2020heidelberg}} & Y & 128-128 & 52.00 \\
 & Heterogeneous LIF\textsubscript{Nat. Commun., 2021\cite{perez2021neural}} & Y & 128 & 60.80 \\
 & 
 % ALIF+multi-Gaussian gradient
 ALIF + GaussinGradient
 \textsubscript{Nat. Mach. Intell., 2021\cite{yin2021accurate}} & Y & 128 & 74.20 \\
 & TC-LIF\textsubscript{AAAI, 2024\cite{zhang2024tclif}} & Y & 128 & 61.09 \\
 & \textbf{DGN(Ours)} & Y & \textbf{128} & \textbf{66.18} \\
 & \textbf{DGN(Ours)} & Y & \textbf{128-128} & \textbf{75.63} \\
\bottomrule
\end{tabular}
\label{tab:sota}
\end{table}

\section{Experiments}

\subsection{Comparison with the State-of-the-Art}

% Speech recognition tasks often contain contexts that are highly correlated over time. Therefore, SNNs are considered to be ideal for such tasks due to their self-recurrent connections. To validate the efficacy of the our DGN model in processing temporal patterns, we conduct comprehensive evaluations on two distinct categories of speech-related datasets. The first category comprises conventional audio classification benchmarks, including Ti46 \cite{ti46} and TIDIGITS \cite{tidigits}. The second category features neuromorphic speech datasets (SHD and SSC) generated through CochleaAMS1b sensor processing, which preserve precise temporal information through event-based encoding. Implementation specifics regarding network architecture and training protocols are detailed in Appendix \ref{app:train_setup}. Our experimental investigation focuses on assessing the performance of multi-layer SNNs implemented with DGN units across all four benchmark datasets: TI46Alpha, TIDIGITS, SHD, and SSC.

% 语音识别任务涉及时间相关的上下文，因此SNN因其自反馈连接特性而被认为是理想选择。为评估我们DGN模型的有效性，我们在两类语音相关数据集上进行了评估：传统音频分类基准（Ti46 \cite{ti46} 和 TIDIGITS \cite{tidigits}）以及通过CochleaAMS1b传感器处理生成的基于事件编码的神经形态语音数据集（SHD和SSC）。网络架构和训练协议的具体细节见附录\ref{app:train_setup}。我们的实验重点评估了四个数据集（TI46Alpha、TIDIGITS、SHD和SSC）上基于DGN单元的多层SNN模型。
Speech recognition tasks involve time-correlated contexts, making SNNs ideal due to their self-recurrent connections. To evaluate the efficiency of our DGN model, we conduct assessments on two categories of speech-related datasets: conventional audio classification benchmarks (Ti46Alpha \cite{ti46} and TIDIGITS \cite{tidigits}) and neuromorphic speech datasets (SHD and SSC) \cite{cramer2020heidelberg}, generated through event-based encoding via CochleaAMS1b sensor processing. Details of the network architecture and training protocols are provided in Appendix \ref{app:train_setup}. Our experiments focus on both feedforward and recurrent SNNs with DGN model across all four datasets.

% 如表1所示，我们提出的生物学启发门控神经元在前馈网络和循环网络中都表现出优异的性能，尽管其使用了更少的神经元和更简化的网络结构，准确率仍然与多个当前SNN领域的SOTA方法相当甚至更优。尤其值得注意的是，在某些任务中，其表现甚至超过了我们基于相同网络结构实现的LSTM模型。这一结果表明，所引入的门控机制有效提升了单个神经元的表达能力，类似于LSTM和GRU对传统RNN的改进效果。
% As shown in Table \ref{tab:sota}, our neuron demonstrates strong performance across both feedforward and recurrent architectures, despite using fewer neurons and simpler network structures. It achieves accuracy on par with or even exceeding several state-of-the-art spiking neural network (SNN) models. Remarkably, on certain tasks, it even outperforms standard LSTM models implemented with the same architecture, highlighting the effectiveness of the proposed biological gating mechanism in enhancing the representational power of individual neurons—akin to how gated RNNs like LSTM and GRU improve over vanilla RNNs.

% 如表1所示，基于动态电导门控机制的DGN模型在多种时序数据处理任务中展现出显著的性能优势。在T146Alpha数据集上，采用递归结构的DGN（Rec=Y）以100节点单隐层架构取得96.31%的准确率，优于同类规模LSTM（96.05%）及多隐层SNN模型（95.90-96.44%）。这种性能增益在TIDIGITS任务中更为显著，递归DGN以99.10%的准确率突破现有记录，较相同架构LSTM（97.88%）提升1.22个百分点。实验结果表明，动态电导机制通过模拟生物神经元膜电导的动态调节特性，有效提升了网络的信息处理效能。该机制赋予单个神经元类似LSTM门控单元的自适应调控能力，但无需引入复杂的分支结构，从而在保持网络拓扑简洁性的同时（如SHD任务中128节点单隐层达85.18%非递归准确率），实现了对传统SNN模型（如TC-LIF的83.08%）和门控RNN模型（如LSTM的89.2%）的双重超越。特别是在SC数据集递归任务中，双隐层DGN以75.63%的准确率超越现有最优SNN方法（ALIF+高斯梯度的74.20%），验证了动态电导机制通过增强神经元内部状态调控能力，可显著提升网络对复杂时序模式的表征深度。这种生物启发的门控机制为构建高效脉冲神经网络提供了新思路，其性能优势在模型规模与计算复杂度严格受限的场景下（如TIDIGITS任务中100节点单隐层架构）体现得尤为显著。
As shown in Tab. \ref{tab:sota}, the feedforward DGN network with a single 100-node hidden layer attains 98.59\% classification accuracy on the TIDIGITS dataset, surpassing comparably structured multilayer spiking neuron networks. Notably, the dual-layer recurrent DGN achieves 75.63\% accuracy on the SSC dataset, outperforming all other approaches. Our proposed DGN show excellent performance in both feedforward and recurrent networks, and their accuracy is comparable to or even better than several current SOTA methods in the field of SNNs, despite using fewer neurons and a simpler network structure. It is worth mentioning that in some tasks, DGN's performance even exceeds that of LSTM model based on the same network structure. These results show that the introduced gating mechanism effectively improves the expressive power of a single neuron. The effectiveness of this mechanism in enhancing the efficiency of neuronal information transmission to process complex time series data has been effectively demonstrated.

\subsection{Overall Performance for Various Perturbation}\label{exp:noise}

% 为了评估所提出的动态门控神经元 (DGN) 模型的鲁棒性，我们构建了一个严格的框架，其中所有模型均在未经人为破坏或噪声处理的原始数据集上进行训练。传统的鲁棒性评估通常使用在训练集和测试集中都包含噪声样本的混合数据集，这可能导致噪声模式记忆和泛化膨胀 \cite{bishop1995training, simard2003best}。该方法引入了显著的数据集构建开销，并且由于模型之间模式学习能力的差异，可能产生误导性的抗噪性评估 \cite{tsipras2018robustness, thams2022evaluating, zhang2021double}。相比之下，我们的方法挑战了具有先前未学习过的噪声模式的模型，从而在次优条件下对其性能进行了更真实的评估。
To evaluate the robustness of the proposed DGN model, we implement a rigorous framework where all models are trained on pristine datasets without artificial corruption or noise. Traditional robustness evaluations typically use hybrid datasets containing noisy samples in both training and testing sets, which can lead to noise pattern memorization. \cite{bishop1995training, simard2003best}. Traditional method introduces significant dataset construction overheads and may yield misleading assessments of noise immunity due to varying pattern learning capacities among models \cite{tsipras2018robustness, thams2022evaluating, zhang2021double}. In contrast, our approach tests models with previously unseen noise patterns, providing a more authentic evaluation of their performance under suboptimal conditions.

\begin{figure}[t]
    \centering
    \includegraphics[width=1\linewidth]
    {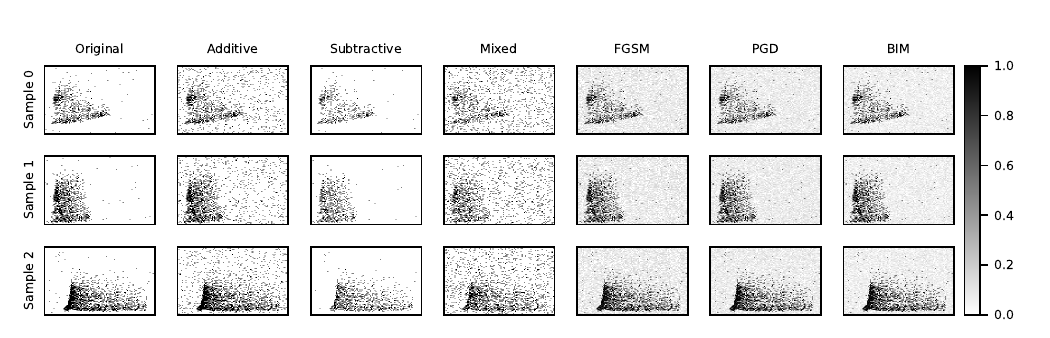}
    \caption{
        Noise sample visualization on SHD dataset. 
        Each row corresponds to one SHD sample, and each column represents a noise type:
        Original, Additive, Subtractive, Mixed, FGSM, PGD, and BIM.
        The horizontal axis indicates time, and the vertical axis represents input channel indices.
    }
    \label{fig:noise}
\end{figure}

\begin{table}[H]
\centering
\small
\caption{
Accuracy (\%) of the proposed DGN under different noise conditions and adversarial attacks on TIDIGITS and SHD. Bold entries indicate the best performance. \textit{HeterLIF} denotes the heterogeneous LIF model proposed by Pérez et al~\cite{perez2021neural}. 
}
\vspace{1mm}
\label{tab:noise_part}
\begin{tabular}{c l c  c c c  c c c}
\toprule
& \multirow{2}{*}{Model} & \multirow{2}{*}{Clean}
& \multicolumn{3}{c}{\textbf{Noise}} 
& \multicolumn{3}{c}{\textbf{Attacks}} \\
& & 
& Additive & Subtractive & Mixed 
& FGSM & PGD & BIM \\
\midrule

\multicolumn{9}{c}{\textbf{ TIDIGITS}} \\
\midrule
% & $\lambda$ &
% & 0.12 & 0.4 & 0.2 
% & 0.006 & 0.003 & 0.006 \\
% \midrule
\multirow{4}{*}{\rotatebox{90}{FF}} 
& LIF           & 97.02     % Clean
                & 46.83     % Addition Noise    0.006
                & 93.70     % Subtractive Noise 0.06
                & 44.20     % Mixed Noise   0.006
                & 39.53     % FGSM Attack   0.003
                & 15.39     % PGD Attack    0.003
                & 15.95 \\  % BIM Attack    0.003
                
& HeterLIF \cite{perez2021neural}      & 96.52     % Clean
                & 77.49     % Addition Noise    0.006
                & 89.37     % Subtractive Noise     0.06
                & 72.78     % Mixed Noise   0.006
                & 52.48     % FGSM Attack   0.003
                & 43.94     % PGD Attack    0.003
                & 43.68 \\  % BIM Attack    0.003
                
& ALIF \cite{bellec2018alif}         & 96.99     % Clean
                & 63.29     % Addition Noise    0.006
                & 93.17     % Subtractive Noise     0.06
                & 60.58     % Mixed Noise   0.006
                & 42.50     % FGSM Attack   0.003
                & 19.80     % PGD Attack    0.003
                & 19.42 \\  % BIM Attack    0.003

& \textbf{DGN}  
                & \textbf{98.59}     % Clean
                & \textbf{95.34}     % Addition Noise   0.006
                & \textbf{93.70}     % Subtractive Noise    0.06
                & \textbf{78.12}     % Mixed Noise  0.006
                & \textbf{90.35}     % FGSM Attack  0.003
                & \textbf{86.76}     % PGD Attack   0.003
                & \textbf{86.88} \\  % BIM Att
\midrule

\multirow{6}{*}{\rotatebox{90}{Rec}} 
& RNN           & 97.09     % Clean
                & 23.64     % Addition Noise
                & 86.76     % Subtractive Noise     0.4
                & 21.66     % Mixed Noise
                & 9.89     % FGSM Attack
                &  0.00     % PGD Attack
                &  0.00 \\  % BIM Attack    0.003
                
& LSTM \cite{hochreiter1997long}         & 97.88     % Clean
                & 65.12     % Addition Noise
                & 79.25     % Subtractive Noise     0.4
                & 64.77     % Mixed Noise
                & 64.97     % FGSM Attack   0.006
                & 60.66     % PGD Attack    0.003
                & 61.01 \\  % BIM Attack    0.003
                
& LIF           & 97.80     % Clean
                & 73.23     % Addition Noise    0.12
                & 89.60     % Subtractive Noise     0.4
                & 67.68     % Mixed Noise
                & 26.55     % FGSM Attack   0.006
                & 61.79     % PGD Attack    0.003
                & 60.70 \\  % BIM Attack    0.003
                
& HeterLIF \cite{perez2021neural}         & 96.29     % Clean
                & 78.97     % Addition Noise    0.12
                & 82.59     % Subtractive Noise     0.4
                & 73.05     % Mixed Noise
                &  8.76     % FGSM Attack   0.006
                & 36.62     % PGD Attack    0.003
                & 35.74 \\  % BIM Attack    0.003
                
& ALIF \cite{bellec2018alif}         & 97.54     % Clean
                & 84.01     % Addition Noise    0.12
                & 86.19     % Subtractive Noise     0.4
                & 79.25     % Mixed Noise
                & 25.04     % FGSM Attack   0.006
                & 62.82     % PGD Attack    0.003
                & 63.18 \\  % BIM Attack    0.003

& \textbf{DGN} 
                & \textbf{99.10}     % Clean
                & \textbf{94.84}     % Addition Noise   0.12
                & \textbf{96.70}     % Subtractive Noise        0.4
                & \textbf{93.86}     % Mixed Noise
                & \textbf{89.40}     % FGSM Attack  0.006
                & \textbf{87.52}     % PGD Attack   0.003
                & \textbf{87.68} \\  % BIM Attack   0.003
\midrule

\multicolumn{9}{c}{\textbf{SHD}} \\
\midrule
% & $\lambda$ &
% & 0.06 & 0.1 & 0.2 
% & 0.006 & 0.003 & 0.006 \\
% \midrule

\multirow{4}{*}{\rotatebox{90}{FF}} 
& LIF           & 77.30     % Clean
                & 29.93     % Addition Noise    0.006
                & 56.32     % Subtractive Noise     0.06
                & 31.44     % Mixed Noise   0.2
                & 51.55     % FGSM Attack   0.006
                & 47.87     % PGD Attack    0.003
                & 47.92 \\  % BIM Attack    0.006
& HeterLIF \cite{perez2021neural}          & 77.77     % Clean
                & 25.49     % Addition Noise    0.06
                & 54.91     % Subtractive Noise     0.1
                & 25.58     % Mixed Noise   0.2
                & 52.23     % FGSM Attack   0.006
                & 50.78     % PGD Attack    0.003
                & 50.89 \\  % BIM Attack    0.006
& ALIF \cite{bellec2018alif}          & 78.02     % Clean
                & 40.25     % Addition Noise    0.06
                & 55.08     % Subtractive Noise     0.1
                & 39.50     % Mixed Noise   0.2
                & 53.31     % FGSM Attack   0.006
                & 51.51     % PGD Attack    0.003
                & 51.57 \\  % BIM Attack    0.006

& \textbf{DGN} 
                & \textbf{85.18}     % Clean
                & \textbf{59.46}     % Addition Noise   0.06
                & \textbf{64.05}     % Subtractive Noise        0.1
                & \textbf{58.87}     % Mixed Noise  0.2
                & \textbf{63.81}     % FGSM Attack  0.006
                & \textbf{61.59}     % PGD Attack   0.003
                & \textbf{61.44} \\  % BIM Attack   0.006
\midrule
\multirow{6}{*}{\rotatebox{90}{Rec}} 
& RNN           & 78.24     % Clean
                & 27.47     % Addition Noise    0.06
                & 52.29     % Subtractive Noise     0.1
                & 28.06     % Mixed Noise   0.2
                & 17.35     % FGSM Attack   0.006
                & 11.93     % PGD Attack    0.003
                & 13.94 \\  % BIM Attack    0.006
                
& LSTM \cite{hochreiter1997long}         & 86.89     % Clean
                & 41.61     % Addition Noise    0.06
                & 64.58     % Subtractive Noise     0.1
                & 39.23     % Mixed Noise   0.6
                & 39.27     % FGSM Attack   0.006
                & 32.01     % PGD Attack    0.003
                & 33.37 \\  % BIM Attack    0.006
                
& LIF           & 75.77     % Clean
                & 9.24     % Addition Noise    0.06
                & 57.44     % Subtractive Noise     0.1
                & 9.25     % Mixed Noise   0.2
                & 17.78     % FGSM Attack   0.006
                & 30.59     % PGD Attack    0.003
                & 31.45 \\  % BIM Attack    0.006
                
& HeterLIF \cite{perez2021neural}         & 79.85     % Clean
                & 39.57     % Addition Noise    0.06
                & 58.19     % Subtractive Noise     0.1
                & 38.87     % Mixed Noise   0.2
                & 44.76     % FGSM Attack   0.006
                & 49.12     % PGD Attack    0.003
                & 49.10 \\  % BIM Attack    0.006
                
& ALIF \cite{bellec2018alif}         & 82.08     % Clean
                & 46.59     % Addition Noise    0.06
                & 63.32     % Subtractive Noise     0.1
                & 47.28     % Mixed Noise   0.2
                & 52.2     % FGSM Attack   0.006
                & 58.01     % PGD Attack    0.003
                & 58.31 \\  % BIM Attack    0.006

& \textbf{DGN}
                & \textbf{87.78}     % Clean
                & \textbf{78.97}     % Addition Noise   0.06
                & \textbf{61.91}     % Subtractive Noise        0.1
                & \textbf{79.35}     % Mixed Noise  0.2
                & \textbf{69.45}     % FGSM Attack  0.006
                & \textbf{66.13}     % PGD Attack   0.003
                & \textbf{66.34} \\  % BIM Attack   0.006
\bottomrule
\end{tabular}
\end{table}

We considered three types of noise commonly encountered in SNNs: additive noise, subtractive noise, and mixed noise. We also evaluated model robustness under three gradient-based adversarial attacks: FGSM\cite{goodfellow2014explainingFGSM}, PGD\cite{madry2018towardsPGD}, and BIM\cite{kurakin2018adversarialBIM}. We conducted anti-noise experiments on the TIDIGITS dataset and the SHD dataset to compare other models with our DGN model. Examples of how different noise types affect the input signals are shown in Fig. \ref{fig:noise}. %The experimental setup is in Appendix \ref{app:noise_setup} and the full experimental data is in Appendix \ref{app:more-exp-res}.

%如表 \ref{tab:noise_part} 所示，基于 DGN 的前馈架构在 TIDIGITS 数据集上加性噪声下的准确率达到 85.62%，比传统的 LIF 模型高出 58.88%，表明其自适应动态电导学习机制能够有效地将噪声特征与信息信号分离。门控架构在噪声抑制方面的优势在各个数据集上都得到了显著体现：在 SHD 的 PGD 攻击下，LSTM 的表现比 vanilla RNN 高出 20.08%，而循环 DGN 的表现比循环 ALIF 高出 8.12%，共同验证了门控机制对噪声鲁棒性的增强。在所有噪声条件和对抗攻击下，DGN 模型都表现出比经典 SNN 神经元更优异的抗干扰能力，保持了最高的基线准确率，并将性能下降降至最低。值得注意的是，它的表现优于传统的 ANN（RNN 和 LSTM）。这些结果证实，受生物启发的门控结构通过动态特征调节引入了鲁棒性改进，其中动态电导机制在对抗弹性中发挥着关键作用。

In Tab. \ref{tab:noise_part}, we select sampling points of different strengths for different perturbation. Noise generation probability $p = 0.006$ for additive noise, and $p = 0.3$ for subtractive noise. The perturbation $\epsilon = 0.003$ for all attacks, and iterative step $k = 4$, step size $\alpha = 0.01$ for PGD, BIM. All results were reproduced by us. Accuracies under the clean condition can be found in Appendix \ref{app:aleanacc}.

As shown in Tab. \ref{tab:noise_part}, the DGN-based feedforward network keeps 95.34\% accuracy under additive noise on the TIDIGITS dataset, surpassing the conventional LIF model by 48.51\%, demonstrating that its adaptive dynamic conductance learning mechanism effectively isolates noise from salient features. The robustness of gated architectures is further evidenced by comparative analyses: under PGD attacks on SHD, LSTM outperforms vanilla RNN by 20.08\%, while the recurrent DGN surpasses recurrent LIF by 35.54\%, collectively validating the robustness enhancement from gated mechanisms.
Across all noise conditions and adversarial attacks, the DGN model exhibits superior resistance compared to classical SNN neurons and conventional ANNs (RNN and LSTM), maintaining the highest baseline accuracy and minimal performance degradation. These results underscore that the biologically inspired gating structure, driven by dynamic conductance modulation, fundamentally enhances robustness.

\subsection{Ablation Study}
\begin{figure}[h]
    \centering
    \includegraphics[width=1.0\linewidth]{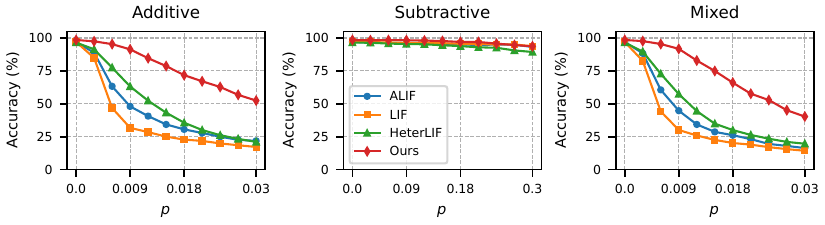}
    \includegraphics[width=1.0\linewidth]{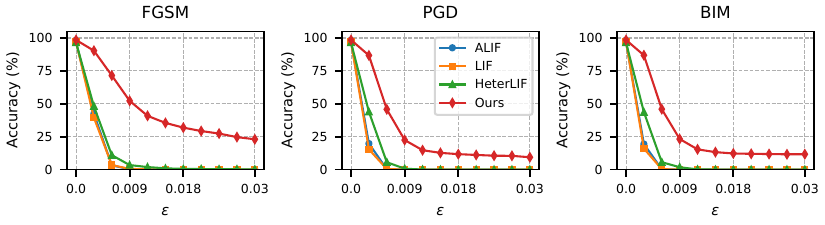}
    \caption{Performance of the model on TIDIGITS using a feedforward network under perturbations of different distribution probabilities \bm{$p$} or attack strengths \bm{$\varepsilon$}.}
    \label{fig:ablation-page1}
\end{figure}

\textbf{Performance under Different Perturbation Strength.} 
We systematically assess the robustness of spiking neuron models by measuring their classification accuracy under escalating perturbation intensities ($p$ or $\varepsilon$). As shown in Fig. \ref{fig:ablation-page1}, the proposed gated neuron model maintains higher classification accuracy with only marginal degradation when subjected to intensified noise disturbances and diverse adversarial attacks. This performance advantage is attributed to the gating mechanism's ability to dynamically adjust the neuron information transmission mode, thereby improving the model's adaptability to perturbations. Extended experiments based on other datasets (Appendix \ref{app:curve}) and structural level analysis (Appendix \ref{app:as_rationality}) further demonstrate the robustness of DGN.

% While it maintains the highest robustness across most perturbation settings, performance declines more noticeably under strong subtractive noise, likely due to excessive input reduction that limits the adaptability of the biologically inspired gating mechanism. 

\section{Conclusion}
% In this paper，为了应对传统SNN研究中缺乏一个具备生物学合理性的门控方案，我们通过重新调研生物学中的电导动态性这一机制，提出了我们的模型，成功通过模拟生物学中的动态电导性质提供了一种生物启发的门控结构，显著的增强了神经元的时空表达能力和信息选择能力，带来了在语音识别任务上的显著提升，同时带来了更强的鲁棒性，面对多种噪声干扰和攻击都显示出相较于传统方法更好的鲁棒性，
% 同时相比传统的基于电导建模的神经元（如HH）展现出更低的能耗，
%为脉冲神经网络的研究提供了一种新的模型方法和研究思路。我们的研究在未来还可以更多的与脉冲神经网络领域的其他方法融合，有着十分丰富的探索空间，同时我们也希望在未来提供更多的带有门控结构的神经元以及时空特性更为丰富的电导模型。
In this work, we address the lack of biologically plausible gating mechanisms in traditional spiking neural networks (SNNs) by revisiting the biophysical principle of dynamic conductance. Inspired by the temporal behavior of biological ion channels, we propose a novel neuron model that implements a biologically inspired gating structure. This mechanism significantly enhances the spatiotemporal expressiveness and information selectivity of the neuron, leading to substantial performance improvements in speech recognition. Moreover, our model demonstrates superior robustness under various noise perturbations and adversarial attacks.
%, and achieves lower energy consumption compared to traditional conductance-based neurons such as Hodgkin-Huxley (HH). 
This work introduces a new modeling paradigm for SNNs, offering insights into both robust computation and biologically grounded design. Future directions include integrating our approach with other advances in the SNNs community, and exploring richer conductance-based gating models with enhanced spatiotemporal properties.

% \section*{Acknowledgment}
\newpage
\bibliography{main}  % references files

\newpage
\appendix

\section{Appendix}

\subsection{Methodology}

\subsubsection{Dynamic Gated Neuron}\label{app:AADN_model}
 % 公式（1）的常微分方程的解析解为：
 Eq.\eqref{1002}\ is:
 \begin{gather}
     \frac{dg_i}{dt} = -\frac{g_i}{\tau_s} + C_i\sum_{t_i^j < t}\delta(t-t_i^j) \label{_1}
 \end{gather}
where $t_i^j$ indicating the arrival time of the $j$th presynaptic spike of the $i$th afferent neuron before time $t$. $\delta(x)$ represents "Dirac delta function", so $\sum_{t_i^j < t}\delta(t-t_i^j)$ is equivalent expression of $z_i(t)$. Then we solve Eq.\eqref{_1} using the general solution method for first-order linear nonhomogeneous differential equations:
\begin{gather}
    \begin{split}
        g_i(T) &= e^{-\int_0^T \frac{1}{\tau_s} dt}(c + \int_0^T{
                e^{\int_0^t \frac{1}{\tau_s} dk} C_i\sum_{t_i^j < t}\delta(t-t_i^j) 
            }dt) \\
            &= e^{-\frac{T}{\tau_s}}(c + C_i\int_0^T{
                  \sum_{t_i^j < t} e^{\frac{t}{\tau_s}} \delta(t-t_i^j)
            }dt) \label{_2}
    \end{split}
\end{gather}

We set $f(t, t_i^j) = e^{\frac{t}{\tau_s}}\delta(t-t_i^j)$, $\Delta t = T/n$, $d = {\left \lceil \frac{t_i^j}{\Delta t} \right \rceil}$(ceiling function). Then:
\begin{gather}
    \begin{split}
        \int_0^T{
                \sum_{t_i^j < t} e^{\frac{t}{\tau_s}} \delta(t-t_i^j)
            }dt
            &= \int_0^T{\sum_{t_i^j < t} f(t, t_i^j)}dt \\
            &= \lim_{n \to \infty} \sum_{k=0}^n \sum_{t_i^j < k\cdot\Delta t} f(k \cdot\Delta t, t_i^j) \Delta t \\
            &= \lim_{n \to \infty} \sum_{t_i^j < T} \sum_{k= d}^n  f(k \cdot\Delta t, t_i^j) \Delta t \\
            &= \sum_{t_i^j < T} \lim_{n \to \infty} \sum_{k= d}^n  f(k \cdot\Delta t, t_i^j) \Delta t \\
            &= \sum_{t_i^j < T} \int_{t_i^j}^T f(t, t_i^j) dt \label{_3}
    \end{split}
\end{gather}
%根据狄拉克的性质我们可以得到$\int_{t_i^j}^T f(t, t_i^j) dt = f(t_i^j, t_i^j) = e^{\frac{t_i^j}{\tau_s}}$. 当0时刻时(t=0)，膜电位$V(t)=0$，带入式子14可以得到c=0。所以，最终得到：
According to the properties of the Dirac delta function, we can get $\int_{t_i^j}^T f(t, t_i^j) dt = f(t_i^j, t_i^j) = e^{\frac{t_i^j}{\tau_s}}$. When $t=0$, the membrane potential $V(t)=0$, and substituting it into Eq.\eqref{_2}, we can get $c=0$. So, we finally get:
\begin{gather}
    \begin{split}
        g_i(T) &= C_i \cdot e^{-\frac{T}{\tau_s}} \sum_{t_i^j < T} e^{\frac{t_i^j}{\tau_s}} \\
            &= C_i \sum_{t_i^j < T} e^{-\frac{T-t_i^j}{\tau_s}} \label{_4}
    \end{split}
\end{gather}
Then, we set $D_i^t = \sum_{t_i^j < t} e^{-\frac{t-t_i^j}{\tau_s}}$, we get Eq.\eqref{221}. In the discrete case, we have Eq.\eqref{223}.
Then:
\begin{equation}
    g_i(t)=C_i\sum_je^{-\frac{t-t_i^j}{\tau_s}}=CiD_i^t \label{1003}
\end{equation}
%将上式带入1式，可得：
Substituting the above formula into Eq.(\ref{1001}), we can get:
\begin{equation}
    \frac{dV}{dt} = -V(g_l + \sum_i^NC_iD_i) + \sum_i^NE_iC_iD_i \label{1004}
\end{equation}
% $E_i$在生物学上认为只有两种取值，但在实际的网络构建过程中，可以设置多种$E_i$值，即每条突触拥有不一定相同的平衡电位。当自由值$E_i$和自由值可训练权重$C_i$相乘时，自然得到另一个可训练权重$W_i$。
In neurobiological computational modeling, classical theoretical frameworks typically posit synaptic equilibrium potential \( E_i \) as a binary-state parameter. However, our network construction process transcend this limitation by permitting heterogeneous equilibrium potential parameters across individual synaptic units. So we set \(E_i\) as a learnable parameter. The mathematical formalization method establishes the synaptic connection weight \(C_i \cdot E_i \) as a learnable parameter \(W_i\) through a multiplicative relationship, because \(C_i \) represents a trainable parameter and \(E_i\) is also a trainable parameter. This parameterization methodology preserves biophysical interpretability while enabling multidimensional regulatory mechanisms for synaptic efficacy. Crucially, such an approach not only transcends the theoretical constraints of conventional bistable equilibrium potentials but also substantially augments the modeling capacity for network dynamics characteristics through the incorporation of continuous-spectrum \( E_i \) values.
So by slightly rearranging Eq.\eqref{1004}, we can get Eq.\eqref{223}-Eq.\eqref{226}.

\subsubsection{Derivation of SDE variance}     \label{app:SDE}
The perturbation satisfies:
\begin{gather}
    \mathbb{E}(\xi(t)) = 0, \xi(t)dt = d\mathbb{W}_t, \mathbb{E}[d\mathbb{W}_t]=0
\end{gather}
In an Itô process\cite{ito1944109}, the following equation. holds:
\begin{gather}
dtdt=0,dtd\mathbb{W}_t=d\mathbb{W}_tdt=0,d\mathbb{W}_td\mathbb{W}_t=dt
\end{gather}
where $\mathbb{W}_t$ is the Brownian motion used to describe random behavior.
We have a linear SDE of a DGN:
\begin{equation}
    \frac{dV}{dt} = -G_0V + \sum W_i \mu_i +\sum \sigma_i \left( W_i - C_i V_{\text{steady}} \right) \xi(t) \label{_5}
\end{equation}
We take the expectation on both sides of dynamic equation Eq.\eqref{_5}. Then we get:
\begin{align}
    \frac{d}{dt}\mathbb{E}[V] 
    &= -G_0\mathbb{E}[V] + \sum W_i \mu_i +\sum \sigma_i \left( W_i - C_i V_{\text{steady}} \right) \mathbb{E}[\xi(t)] \\
    &= -G_0\mathbb{E}[V] + \sum W_i \mu_i   \label{_6}
\end{align}
In steady state:
\begin{gather}
    \mathbb{E}[V] = \frac{\sum W_i \mu_i}{G_0}  \label{_7}
\end{gather}
Applying  Itô calculus\cite{ito1944109} to $V^2$, we can get the calculation formula:
\begin{gather}
    d(V^2) = 2VdV+(dV)^2 \label{_8}
\end{gather}
Substituting Eq.\eqref{_5} into the above equation, we get
\begin{gather}
    d(V^2) =
    -2G_0V^2dt+(\sum2W_i\mu_i)Vdt
    +\left[\sum\sigma_i(W_i-C_iV_{\text{steady}})\right]^2dt\\
    +2\left[\sum\sigma_i(W_i-C_iV_{\text{steady}})\right]V\xi(t)dt \label{_9}
\end{gather}
Taking the expectation on both sides, we get:
\begin{gather}
    \frac{d\mathbb{E}[V^2]}{dt} = 
    -2G_0\mathbb{E}[V^2]+(\sum2W_i\mu_i)\mathbb{E}[V]
    +\left[\sum\sigma_i(W_i-C_iV_{\text{steady}})\right]^2\\
    +2\left[\sum\sigma_i(W_i-C_iV_{\text{steady}})\right]V\mathbb{E}[\xi(t)] \label{_10}
\end{gather}
Substituting Eq.\eqref{_7} into the above equation, we can get the following when taking steady state:
\begin{gather}
    \langle V^2 \rangle = \mathbb{E}[V^2]-(\mathbb{E}[V])^2=\frac{\left[\sum_{i=1}^N \sigma_i \left( W_i - \frac{C_i \sum_{j=1}^N W_j \mu_j}{G_0} \right)\right]^2}{2G_0} \label{_11}
\end{gather}
Similarly, the steady-state variance of LIF neurons can be obtained.

\subsubsection{Training DGN-SNNs with BPTT}   \label{app:bptt}
\begin{figure}
    \centering
    \includegraphics[width=0.4\linewidth]{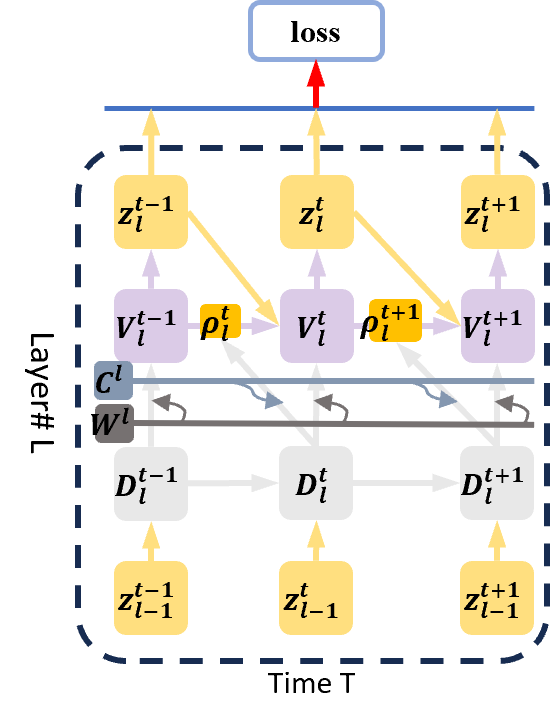}
    \caption{DGN unfolds over three time steps}
    \label{fig:BPTT}
\end{figure}

The network outputs at each timestep $t$ are given by $o_t = W_L z_L^t$. Classification is based on the average of these outputs across all timesteps, computed as $y_{pred} = \frac{1}{T}\sum_{t=1}^T o_t$. The loss function $\mathcal{L}$ is defined over averaged outputs and is typically formulated as $E = \ell(y_{pred}, y)$, where $y$ represents the true labels and $\ell$ could be the cross-entropy function, as noted in various studies\cite{zheng2021going, meng2023towards, fang2023spikingjelly, wang2023adaptive}

BPTT unfolds the iterations described in Eq.\ref{225}, and propagates gradients back along the computational graphs across both temporal and spatial dimensions, as illustrated in Fig. \ref{fig:BPTT}. Subsequently, the weight update for single layer is determined among all timesteps $T$:
\begin{align}
    \frac{dE}{dW_i} = &\sum_t^T \frac{dE}{dz^t} \frac{dz^t}{dW_i} \label{10064} \\
    \frac{dE}{dC_i} = &\sum_t^T \frac{dE}{dz^t} \frac{dz^t}{dC_i} \label{10065} \\
     \frac{dE}{dz^t} = \frac{\partial E}{\partial z^t} + &\sum_{k=t+1}^T[\prod_{j=t+1}^k(-\vartheta\Psi^j)\frac{\partial E}{\partial z^k}] \label{10066} \\
    \frac{dz^t}{dW_i} = \Psi^t\{D_i^t + &\sum_{k=1}^{t-1}[\prod_{j=k+1}^{t}(\rho^j-\vartheta\Psi^{j-1})D_i^k]\} \label{10067} \\
     \frac{dz^t}{dC_i} = \Psi^t\{-f'V^{t-1}D_i^t + &\sum_{k=1}^{t-1}[\prod_{j=k+1}^{t}(\rho^j-\vartheta\Psi^{j-1})(-f'V^{k-1}D_i^k)]\} \label{10068}
\end{align}
where:
\begin{itemize}
    \item $\Psi^t$: surrogate gradient, $\Psi^t = dz^t/dV^t$
    \item $F'$: derivative of the truncated function $\phi$ is Eq.\eqref{224}
    \item $f'$: the value of $F'$ at $1-(g_l+\sum_i^NC_iD_i^t)$, i.e. $f'=F'(1-(g_l+\sum_i^NC_iD_i^t))$
\end{itemize}
In the process of gradient propagation using BPTT, it is also necessary to manually set surrogate function to calculate surrogate gradient $\Psi^t$, which are used as $dz^t/dV^t$, that is:
\begin{gather}
    \Psi^t = \frac{dz^t}{dV^t} \label{10038}
\end{gather}

The detailed derivation process of Eq.\eqref{10064} $\sim$ Eq.\eqref{10068} is as follows. The gradient of the loss function $E$ with respect to the trainable weights $C_i$ and $W_i$ of synapse $i$ is:
\begin{align}
    \frac{dE}{dW_i} = \sum_t^T \frac{dE}{dz^t} \frac{dz^t}{dW_i} \label{10039}\\
    \frac{dE}{dC_i} = \sum_t^T \frac{dE}{dz^t} \frac{dz^t}{dC_i} \label{10040}
\end{align}

Combining the calculation graph, we can obtain
\begin{equation}
    \frac{dE}{dz^t} = \frac{\partial E}{\partial z^t} + \frac{dE}{dz^{t+1}}\frac{dz^{t+1}}{dz^t} \label{10041}
\end{equation}

Then:
\begin{equation}
    \frac{dz^{t+1}}{dz^t} = \frac{dz^{t+1}}{dV^{t+1}}\frac{\partial V^{t+1}}{\partial z^t} \label{10042}
\end{equation}

According to  Eq.\eqref{226} and Eq.\eqref{10038}, we obtain respectively:
\begin{equation}
    \frac{\partial V^{t+1}}{\partial z^{t}} = -\vartheta \label{10043}\\
\end{equation}
\begin{equation}
    \frac{dz^{t+1}}{dV^{t+1}} = \Psi^{t+1} \label{10044}\\
\end{equation}

By combining the above formula and substituting Eq.\eqref{10042} into Eq.\eqref{10041}, we obtain:
\begin{equation}
     \frac{dE}{dz^t} = \frac{\partial E}{\partial z^t} - \vartheta\Psi^{t+1}\frac{dE}{dz^{t+1}} \label{10045}
\end{equation}

To carry out the analysis, for any time $1 \le t \le T$, we expand the recursion:
\begin{equation}
     \frac{dE}{dz^t} = \frac{\partial E}{\partial z^t} + \sum_{k=t+1}^T[\prod_{j=t+1}^k(-\vartheta\Psi^j)\frac{\partial E}{\partial z^k}] \label{10046}
\end{equation}

According to Eq.\eqref{10039}Eq.\eqref{10040} combined with the calculation graph, we get:
\begin{equation}
    \frac{dz_t}{dW_i} = \frac{dz^t}{dV^t} \frac{dV^t}{dW_i} \label{10047}
\end{equation}
\begin{equation}
    \frac{dz_t}{dC_i} = \frac{dz^t}{dV^t} \frac{dV^t}{dC_i} \label{10048}
\end{equation}

According to the calculation diagram of DGN over time, combined with formula \eqref{225}, we can get:
\begin{equation}
    \frac{dV^t}{dW_i} = \frac{\partial V^t}{\partial W_i} + \frac{\partial V^t}{\partial V^{t-1}}\frac{dV^{t-1}}{dW_i} + \frac{\partial V^t}{\partial z_{t-1}}\frac{dz^{t-1}}{dW_i} \label{10049}
\end{equation}
\begin{equation}
    \frac{dV_t}{dC_i} = \frac{\partial V^t}{\partial \rho^t}\frac{d\rho^t}{dC_i} +  \frac{\partial V^t}{\partial V^{t-1}}\frac{dV^{t-1}}{dC_i} + \frac{\partial V^t}{\partial z^{t-1}}\frac{dz^{t-1}}{dC_i} \label{10050}
\end{equation}

Substitute Eq.\eqref{10047} into Eq.\eqref{10049}, substitute Eq.\eqref{10048} into Eq.\eqref{10050}, and arrange them to get:
\begin{equation}
    \begin{split}
        \frac{dV^t}{dW_i} 
            &= \frac{\partial V^t}{\partial W_i} + \frac{\partial V^t}{\partial V_{t-1}}\frac{dV^{t-1}}{dW_i} + \frac{\partial V^t}{\partial z_{t-1}}\frac{dz^{t-1}}{dV^{t-1}}\frac{dV^{t-1}}{dW_i} \\
            &= \frac{\partial V^t}{\partial W_i} + ( \frac{\partial V^t}{\partial V_{t-1}} + \frac{\partial V^t}{\partial z_{t-1}}\frac{dz^{t-1}}{dV^{t-1}} )\frac{dV^{t-1}}{dW_i} \\ \label{10051}
    \end{split}
\end{equation}
\begin{equation}
    \begin{split}
        \frac{dV^t}{dC_i} 
            &= \frac{\partial V^t}{\partial \rho^t}\frac{d\rho^t}{dC_i} +  \frac{\partial V^t}{\partial V^{t-1}}\frac{dV^{t-1}}{dC_i} + \frac{\partial V^t}{\partial z^{t-1}}\frac{dz^{t-1}}{dV^{t-1}} \frac{dV^{t-1}}{dC_i} \\
            &= \frac{\partial V^t}{\partial \rho^t}\frac{d\rho^t}{dC_i} +  (\frac{\partial V^t}{\partial V^{t-1}} + \frac{\partial V^t}{\partial z^{t-1}}\frac{dz^{t-1}}{dV^{t-1}}) \frac{dV^{t-1}}{dC_i}  \\ \label{10052}
    \end{split}
\end{equation}

According to the Eq.\eqref{223}$\sim$\eqref{226}, we get:
\begin{align}
    \frac{\partial V^t}{W_i} &= D_i^t \label{10053}\\
    \frac{\partial V^t}{\partial V_{t-1}} &= \rho^t \label{10054}\\
    \frac{\partial V^t}{\partial z^{t-1}} &= -\vartheta \label{10055}\\
    \frac{dz^{t}}{dV^{t}} &= \Psi^{t} \label{10056} \\
    \frac{dz^{t-1}}{dV^{t-1}} &= \Psi^{t-1} \label{10057}\\
    \frac{\partial V^t}{\rho_t} &= V^{t-1} \label{10058}\\
    \frac{d\rho^t}{dC_i} = -F'(1-&(g_l+\sum_i^NC_iD_i^t))D_i^t \label{10059}
\end{align}

Substitute the above formula into Eq.\eqref{10051}, Eq.\eqref{10052} and sort it out to get:
\begin{equation}
    \frac{dV^t}{dW_i} = (\rho^t-\vartheta\Psi^{t-1})\frac{dV^{t-1}}{dW_i} + D_i^t \label{10060}
\end{equation}
\begin{equation}
    \frac{dV^t}{dC_i} = (\rho^t-\vartheta\Psi^{t-1})\frac{dV^{t-1}}{dC_i} - f'V^{t-1}D_i^t \label{10061}
\end{equation}

Expand the recursive calculation of equations Eq.\eqref{10060} and Eq.\eqref{10061}, and we get Eq.\eqref{10064} $\sim$ Eq.\eqref{10068}

\subsection{Experiments}

\subsubsection{Datasets}

\begin{table}[h]
\centering
\caption{Network parameters for different datasets.}
\vspace{1mm}
\begin{tabular}{llcccc}
\toprule
\textbf{Dataset} & \textbf{Network} & $\tau_m$ & $\tau_s$ & $\vartheta$  & 
 ($c$, $w$)\\
\midrule
\multirow{2}{*}{Ti46Alpha} 
    & feedforward  & 10.00 & 2.0 & 1.00 & (0.01 ± 0.005, 0.01 ± 0.005) \\
    & recurrent   & 15.00 & 1.50 & 1.00 &  (0.01 ± 0.005, 0.01 ± 0.005) \\
\midrule
\multirow{2}{*}{TIDIGITS} 
    & feedforward  & 100.00 & 1.0 & 1.00 & (0.01 ± 0.005, 0.001 ± 0.0005) \\
    & recurrent  0 & 10.00 & 2.50 & 1.00  & (0.01 ± 0.005, 0.01 ± 0.005) \\
\midrule
\multirow{2}{*}{SHD} 
    & feedforward  & 1.00 & 0.02 & 1.00 & (0.01 ± 0.005, 0.01 ± 0.005) \\
    & recurrent   & 1.00 & 0.02 & 1.00  &  (0.001 ± 0.0005, 0.001 ± 0.0005)\\
\midrule
\multirow{2}{*}{SSC} 
    & feedforward  & 1.00 & 0.02 & 1.00 & (0.01 ± 0.005, 0.01 ± 0.005) \\
    & recurrent    & 1.00 & 0.02 & 1.00  &  (0.01 ± 0.005, 0.01 ± 0.005)\\
\bottomrule
\end{tabular}
\label{tab:network-params}
\end{table}

\textbf{TI46Alpha}: TI46Alpha is the full alphabets subset of the TI46 Speech corpus \cite{ti46} and contains spoken English alphabets from 16 speakers. There are 4,142 and 6,628 spoken English examples in 26 classes for training and testing, respectively. The threshold encoding mechanism \cite{gutig2009time} is used to encode the audio information into spike pattern. First, a spectrogram is computed  with consecutive Fourier transforms (legacy function) from the original sound wave. Then the spectrogram is filtered by a filter bank of 16 triangular filters to obtain a mel-scale spectrogram. Next, for each mel-scale spectrogram bin corresponding to a filter, 30 neurons are used to encode its energy changes as spikes. Thus, a total of 480 neurons are used to encode an audio sample (more details, see \cite{gutig2009time}). In order to increase the generalization ability of the model, we added 20 empty channels, each original audio has been converted into spike trains over 500 input channels.

% \textbf{TI46-Digit}: TI46-Digits is the full digits subset of the TI46 Speech corpus (Liberman et al. 1991). It contains 1,594 training examples and 2,542 testing examples of 10 utterances for each of digits ”0” to ”9” spoken by 16 different speakers. The same preprocessing used for TI46-Alpha is adopted.

\textbf{TIDIGITS}:  TIDIGITS is a widely used speech recognition dataset that contains the utterances of 11 words from the digits “zero” to “nine” and “oh.” It contains a training set of 2464 samples and a test set of 2486 samples. The same preprocessing used for TI46Alpha is adopted.

\textbf{SHD}: The Spiking Heidelberg Digits dataset is a spike based sequence classification benchmark, consisting of spoken digits from 0 to 9 in both English and German (20 classes). The dataset contains recordings from twelve different speakers, with two of them only appearing in the test set. Each original waveform has been converted into spike trains over 700 input channels. The train set contains 8,332 examples, and the test set consists of 2,088 examples (no validation set). In our experiments, we reduce the time resolution to speed up the simulation. Therefore, the preprocessed samples only have about 250 time steps. We determine that a channel has a spike at a certain time step of the preprocessed sample if there’s at least one spike among the corresponding several time steps of the original sample.

\textbf{SSC}: The Spiking Speech Command dataset, another spike-based sequence classification benchmark, is derived from the Google Speech Commands version 2 dataset and contains 35 classes from a large number of speakers. The original waveforms have been converted to spike trains over 700 input channels. The dataset is divided into train, validation, and test splits, with 75,466, 9,981, and 20,382 examples, respectively. The same preprocessing used for SHD is adopted.

% \textbf{N-TIDIGITS}: The N-TIDIGITS \cite{Anumula2018FeatureRF} is the neuromorphic version of the well-known speech dataset TIDIGITS \cite{tidigits}, and consists of recorded spike responses of a 64-channel CochleaAMS1b sensor in response to the original audio waveforms. The dataset includes both single digits and connected digit sequences with a vocabulary consisting of 11 digits including “oh,” “zero” and the digits “1” to “9”. There are 55 male and 56 female speakers with 2,475 single digit examples for training and the same number of examples for testing. In the original dataset, each sample lasts about 0.9s. The same preprocessing used for SHD is adopted.

\subsubsection{Training Setup} \label{app:train_setup}

We train the Ti46Alpha and TIDIGITS datasets for 64 epochs utilizing the Adam optimizer. Their learning rate are set to 0.001 for both feedforward and recurrent networks. For SHD and SSC datasets, we train the models for 128 epochs using the Adam optimizer. Their learning rate are set to 0.001 as well. Unlike standard binary spike trains, the SHD dataset have been temporally preprocessed to aggregate spikes within 4ms-windows\cite{zhang2024tclif, yin2021accurate, cramer2020heidelberg}, resulting in integer spike counts per time step. We train all of the datasets on Nvidia GeForce RTX 4060 GPUs with 8GB memory for feedforward network and Nvidia GeForce RTX 4090 GPUs with 24GB memory for recurrent network.

To clearly assess the contribution of our proposed neuron model, we intentionally use simple fully connected architectures\cite{zhang2024tclif} in all experiments. This choice minimizes interference from other architectural components and ensures that any performance gain arises from the neuron dynamics themselves. We outline the specific hyperparameter settings for the all neuron model(both our DGN and other method we reproduced) in Tab. \ref{tab:network-params}, encompassing the time constant for membrane( $\tau_m$ ) and synapse( $\tau_s$ ), and spike threshold( $\vartheta$ ), and the ($c$, $w$) is the initial value for trainable weights($C$, $W$), where c in only use in DGN model.

\subsubsection{Accuracies Under the Clean Condition} \label{app:aleanacc}
% 我们根据所引用论文复现的数据集结果如表\ref{tab:cleanacc}所示，为了保证后续噪声和对抗攻击实验的公平性，我们均使用了同样的超参数设置来获取基础模型，这可能是导致我们的复现结果相比所引用论文提供的准确率有一定的损失的原因之一。
We reproduce the results on the datasets following the referenced paper, as shown in Tab. \ref{tab:cleanacc}. To ensure fairness in the subsequent experiments involving noise and adversarial attacks, we use the same hyperparameter settings across all runs to obtain the base models. This consistent setup may partially explain the discrepancy between our reproduced accuracies and those reported in the original paper.

For Ti46Alpha and TIDIGITS datasets, we use 100 neuron with single layer for hidden layer, and for SHD and SSC neuron, we use 128 neuron with single layer for hidden layer.

We conducted three runs of DGN model using the same initialization but different random seeds. In Tab. \ref{tab:cleanacc}, the bottom row presents the mean and standard deviation, while the second-to-last row reports the best result among the three runs. The best-performing model was subsequently used for the noise robustness experiments. This parameter configuration, when applied to other models, might not fully optimize their performance under clean conditions, potentially resulting in varying comparative advantages across different models.

% Notably, only our model is evaluated with multiple runs; for all reproduced model, we report only the best result, that's why the clean accuracy for other models may not benefit as much from the additional component, which results in HeterLIF and ALIF performing worse than LIF in some datasets.

% Notably, to ensure a fair comparison with other models, we use the same parameter settings to train all models. Compared with the results reported in the reference, due to differences in the running machines and different hyperparameter selections, the slight differences in the classification accuracy of the reproduced results are reasonably acceptable.

% Table generated by Excel2LaTeX from sheet 'Sheet1'

\subsubsection{Noise Setup}\label{app:noise_setup}

\textbf{Additive}     \\
Each element in the input tensor is independently perturbed by adding a random binary value sampled from a Bernoulli distribution with probability $p$. The noise tensor has the same shape as the input. After the addition, the resulting values are clamped from below to ensure that no negative values remain.

% \begin{algorithm}
% \caption{Additive Noise}
% \begin{algorithmic}[1]
% \Require Data tensor $data$; Probability $p$
% \Ensure Perturbed data with Bernoulli noise
% \State $noise \gets \text{Bernoulli}(p)$ samples with same shape as $data$
% \State $perturbed\_data \gets data + noise$
% \State $perturbed\_data \gets \max(perturbed\_data, 0)$ 
% %\Comment{Clamp negative values to 0}
% \State \Return $perturbed\_data$
% \end{algorithmic}
% \end{algorithm}

\textbf{Subtractive} \\
Each non-zero element in the input tensor is independently perturbed by subtracting a random binary value sampled from a Bernoulli distribution with probability $p$. The perturbation only occurs where the original data is greater than zero. After the subtraction, the resulting values are clamped from below to ensure no negative values remain.

% \begin{algorithm}
% \caption{Subtractive Noise}
% \begin{algorithmic}[1]
% \Require Data tensor $data$; Probability $p$
% \Ensure Perturbed data with Bernoulli noise
% \State $noise \gets \text{Bernoulli}(p)$ samples with same shape as $data$
% \State $mask \gets (data > 0)$ 
% %\Comment{Only perturb non-zero entries}
% \State $perturbed\_data \gets data - noise \times mask$
% \State $perturbed\_data \gets \max(perturbed\_data, 0)$ 
% %\Comment{Clamp negative values to 0}
% \State \Return $perturbed\_data$
% \end{algorithmic}
% \end{algorithm}

\textbf{Mixed} \\
This approach combines both additive and subtractive Bernoulli noise. For non-zero elements, noise is subtracted with a higher probability scaled by a factor (default 10×). For zero elements, noise is added with the original probability $p$. All perturbations are performed independently, and the result is clamped to ensure no negative values remain. Since the input non-zero valid data is very sparse, only when the probability of subtractive noise is high can the interference effect be equal to (or even lower than) that of additive noise. Therefore, when constructing mixed noise, the probability of subtractive noise is magnified by 10 times.

% \begin{algorithm}
% \caption{Mixed Noise}
% \begin{algorithmic}[1]
% \Require Data tensor $data$; Probability $p$; Deletion scale factor $\gamma$ (default: 10)
% \Ensure Perturbed data with mixed Bernoulli noise
% \State $delete\_mask \gets (data > 0)$ 
% %\Comment{Only delete from non-zero entries}
% \State $delete\_p \gets \min(p \times \gamma, 1)$ 
% %\Comment{Scale and cap the deletion probability}
% \State $delete\_noise \gets \text{Bernoulli}(delete\_p)$
% \State $add\_mask \gets (data == 0)$ 
% %\Comment{Only add to zero entries}
% \State $add\_noise \gets \text{Bernoulli}(p)$
% \State $perturbed\_data \gets data - delete\_mask \times delete\_noise + add\_mask \times add\_noise$
% \State $perturbed\_data \gets \max(perturbed\_data, 0)$ 
% %\Comment{Clamp negative values to 0}
% \State \Return $perturbed\_data$
% \end{algorithmic}
% \end{algorithm}

Given a classification model $f$ with dataset $(x, y_{\text{true}})$, where $x$ is the clean image and $y_{\text{true}}$ is the corresponding correct label. The formulations of the attacks we used in this study are described as follows:

\textbf{FGSM}      \\
FGSM aims to perturb the original data $x$ along the sign direction of the gradient on loss function with one step to increase the perturbed linear output, thus fool the network, it can be formalized as follows:
\begin{equation}
    \hat{x} = x + \epsilon \cdot \text{sign}(\nabla_x \mathcal{L}(f(x), y_{\text{true}})),
\end{equation}
where $\text{sign}(\cdot)$ is an odd mathematical function that extracts the sign of a real number.

\textbf{PGD}       \\
PGD attack is the iterative variant of FGSM. It first starts from a random perturbation in the $L_p$-norm constraint around the original sample $x$, then takes a gradient iteration step in the sign direction to achieve the greatest loss output, it can be formalized as follows:
\begin{equation}
    \hat{x}^0 = x + \mathcal{U}(-\epsilon, +\epsilon),
\end{equation}
\begin{equation}
    \hat{x}^{k+1} = \text{Clip}_{x,\epsilon} \left\{ \hat{x}^k + \alpha \cdot \text{sign}(\nabla_{\hat{x}^k} \mathcal{L}(f(\hat{x}^k), y_{\text{true}})) \right\},
\end{equation}
where $k$ is the iterative step, $\alpha$ is step size for each attack iteration, $\epsilon$ controls the perturbation level. $\mathcal{U}(\cdot)$ is a uniform function, $\text{Clip}_{x,\epsilon}\{\cdot\}$ is the function which performs per-pixel clipping of the image $\hat{x}$, so the result will be in $L_\infty$-norm $\epsilon$-neighborhood of the original image $x$.

\textbf{BIM}       \\
Both BIM and PGD attacks are iterative attacks. Different from PGD attacks, BIM updates the adversarial samples starting from the original image.

% Table generated by Excel2LaTeX from sheet 'Sheet1'
% Table generated by Excel2LaTeX from sheet 'Sheet1'
\begin{table}[htbp]
  \centering
  \caption{Accuracy of each method we reproduced on different datasets without noise or attacks.}
    \begin{tabular}{cccccccccc}
    \toprule
    % \multicolumn{10}{c}{\textbf{Clean Accuracy(\%)}} \\
    % % \midrule
    \multicolumn{2}{c}{Method} & \multicolumn{2}{c}{Ti46alpha} & \multicolumn{2}{c}{Tidigits} & \multicolumn{2}{c}{SHD} & \multicolumn{2}{c}{SSC} \\
    \midrule
    \multirow{5}[2]{*}{FF} & LIF   & \multicolumn{2}{c}{94} & \multicolumn{2}{c}{97.02} & \multicolumn{2}{c}{77.3} & \multicolumn{2}{c}{47.72} \\
          & HeterLIF\cite{perez2021neural} & \multicolumn{2}{c}{93.5} & \multicolumn{2}{c}{96.52} & \multicolumn{2}{c}{76.76} & \multicolumn{2}{c}{55.59} \\
          & ALIF\cite{bellec2018alif}  & \multicolumn{2}{c}{93.85} & \multicolumn{2}{c}{96.99} & \multicolumn{2}{c}{78.02} & \multicolumn{2}{c}{49.17} \\
          & \multirow{2}[1]{*}{DGN} & \multicolumn{2}{c}{95.69} & \multicolumn{2}{c}{98.91} & \multicolumn{2}{c}{85.18} & \multicolumn{2}{c}{58.77} \\
          &       & \multicolumn{2}{c}{(95.60 ± 0.08)} & \multicolumn{2}{c}{(98.24 ± 0.34)} & \multicolumn{2}{c}{(84.6 ± 0.42)} & \multicolumn{2}{c}{(58.34 ± 0.06)} \\
    \midrule
    \multirow{7}[2]{*}{Rec} & RNN   & \multicolumn{2}{c}{91.89} & \multicolumn{2}{c}{97.09} & \multicolumn{2}{c}{78.24} & \multicolumn{2}{c}{72.91} \\
          & LSTM\cite{hochreiter1997long}  & \multicolumn{2}{c}{96.05} & \multicolumn{2}{c}{97.88} & \multicolumn{2}{c}{86.89} & \multicolumn{2}{c}{75.95} \\
          & LIF   & \multicolumn{2}{c}{90.89} & \multicolumn{2}{c}{97.8} & \multicolumn{2}{c}{75.77} & \multicolumn{2}{c}{53.16} \\
          & HeterLIF\cite{perez2021neural} & \multicolumn{2}{c}{91.31} & \multicolumn{2}{c}{96.29} & \multicolumn{2}{c}{79.85} & \multicolumn{2}{c}{63.63} \\
          & ALIF\cite{bellec2018alif}  & \multicolumn{2}{c}{90.28} & \multicolumn{2}{c}{97.54} & \multicolumn{2}{c}{82.08} & \multicolumn{2}{c}{55.96} \\
          & \multirow{2}[1]{*}{DGN} & \multicolumn{2}{c}{96.31} & \multicolumn{2}{c}{99.1} & \multicolumn{2}{c}{87.78} & \multicolumn{2}{c}{66.18} \\
          &       & \multicolumn{2}{c}{(95.74 ± 0.35)} & \multicolumn{2}{c}{(98.67 ± 0.06)} & \multicolumn{2}{c}{(86.33 ± 0.58)} & \multicolumn{2}{c}{(65.72 ± 0.27)} \\
    \bottomrule
    \end{tabular}%
  \label{tab:cleanacc}%
\end{table}%

\subsection{Ablation Study}
\subsubsection{Performance under Different Perturbation Strength} \label{app:curve}

We systematically evaluate the robustness of multiple spiking neuron models on TIDIGITS and SHD datasets using both feedforward and recurrent networks by quantifying the performance loss under gradually increasing parameter perturbations(\bm{$p$} or \bm{$\varepsilon$}). % in Fig.(\ref{fig:tidigits_rec})-Fig.(\ref{fig:shd_rec}). 

\begin{figure}[h]
    \centering
    \includegraphics[width=1.0\linewidth]{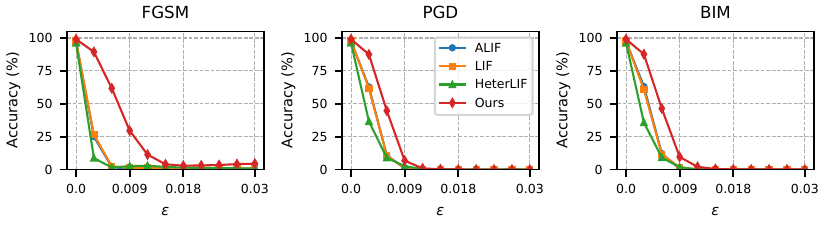}
    \includegraphics[width=1.0\linewidth]{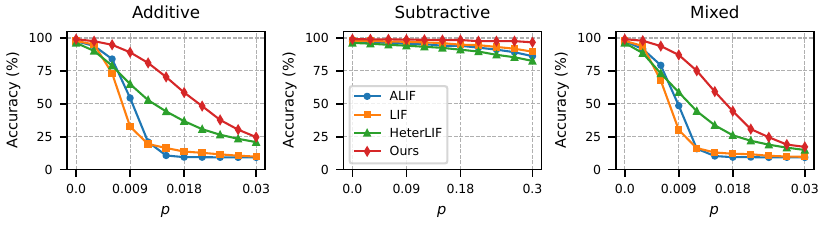}
    \caption{Performance of the model on TIDIGITS using a recurrent network under perturbations of different distribution probabilities \bm{$p$} or attack strengths \bm{$\varepsilon$}.}
    \label{fig:tidigits_rec}
\end{figure}

\begin{figure}[h]
    \centering
    \includegraphics[width=1.0\linewidth]{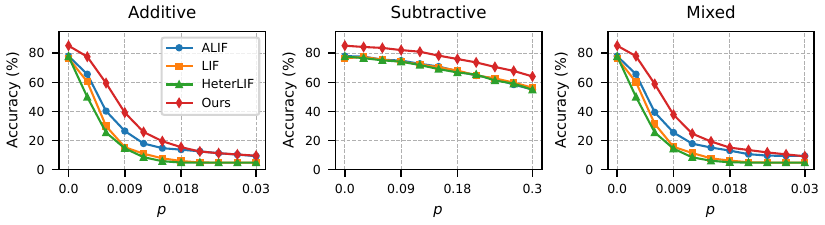}
    \includegraphics[width=1.0\linewidth]{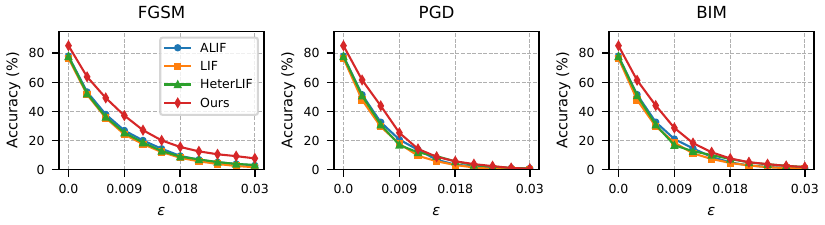}
    \caption{Performance of the model on SHD using a feedforward network under perturbations of different distribution probabilities \bm{$p$} or attack strengths \bm{$\varepsilon$}.}
    \label{fig:shd_ff}
\end{figure}

\begin{figure}[h]
    \centering
    \includegraphics[width=1.0\linewidth]{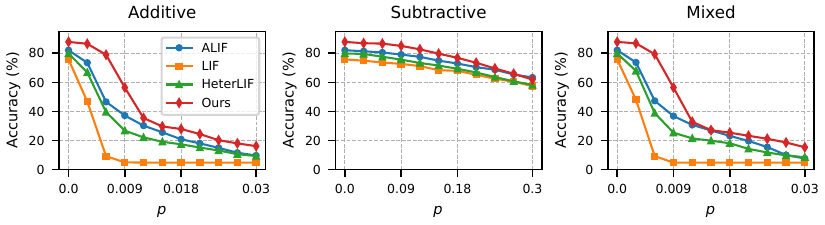}
    \includegraphics[width=1.0\linewidth]{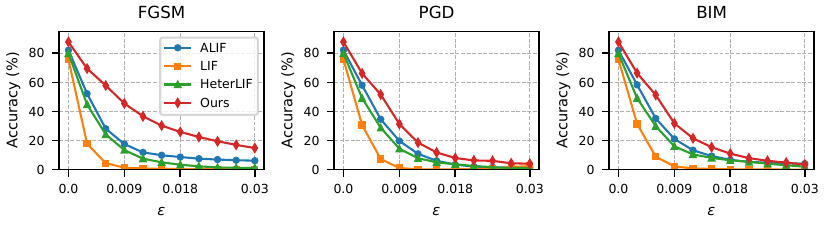}
    \caption{Performance of the model on SHD using a recurrent network under perturbations of different distribution probabilities \bm{$p$} or attack strengths \bm{$\varepsilon$}.}
    \label{fig:shd_rec}
\end{figure}

As illustrated in Fig. \ref{fig:tidigits_rec} - Fig. \ref{fig:shd_rec}, our neuron model outperforms others in terms of accuracy under increasing perturbation intensities. It consistently maintains the highest accuracy and exhibits the lowest degradation across within a reasonable perturbation range. %However, under subtractive noise conditions, our model does not outperform others because the input information is overly reduced, limiting the adaptability of our input-driven biological gates.

% As illustrated in Fig.(\ref{fig:ff_attack_curve}), our gated neuron exhibits consistently superior accuracy and minimal performance degradation across a range of adversarial attack types with increasing perturbation strengths.

These results demonstrate that our neuron model is able to filter out interfering information while maintaining excellent effective information transfer efficiency, highlighting the effectiveness of the proposed bio-inspired gating mechanism in enhancing the model's robustness to a variety of perturbation patterns.

\subsubsection{Rationality of Gated Structure}\label{app:as_rationality}
% 从1式中可以观察看到，我们的门结构是由两部分构成的：其中静态结构和可训练权重无关，在网络前向传播过程中保持不变；动态结构和可训练权重$C_i$绑定，在网络前向传播过程中始终保持着动态变化，并可通过梯度学习算法学习数据集特性。
As demonstrated in Eq.\eqref{431}, the proposed gating mechanism exhibits a dual-component architecture comprising static and dynamic elements. The static component, represented by the leakage conductance $g_l$, remains invariant to trainable parameters and preserves constant characteristics throughout network propagation. In contrast, the dynamic component $\sum_i^NC_iD_i$ establishes parametric dependence on learnable weights $\{C_i\}$, enabling continuous adaptation during forward propagation through gradient-based optimization. 
\begin{gather}
    \frac{dV}{dt} = -(\underbrace{g_l}_{\text{static}} + \underbrace{\textstyle \sum_i^NC_iD_i}_{\text{dynamic}})V + \sum_iW_iD_i \label{431}
\end{gather}

\begin{table}[htbp]
\caption{Ablation Study for Gate Structure}
\centering
\begin{tabular}{c l c  c c c  c c c}
\toprule
& \multirow{2}{*}{Model} & \multirow{2}{*}{Clean}
& \multicolumn{3}{c}{\textbf{Noise}} 
& \multicolumn{3}{c}{\textbf{Attacks}} \\
& & 
& Addition & Subtractive & Mixed 
& FGSM & PGD & BIM \\
\midrule
\multirow{3}{*}{\rotatebox{90}{FF}} 
% & LIF               & xx.xx 
%                     & xx.xx 
%                     & xx.xx 
%                     & xx.xx 
%                     & xx.xx 
%                     & xx.xx 
%                     & xx.xx \\
& w$\backslash$o D  & 97.02
                    & 46.83 
                    & 93.70 
                    & 44.20 
                    & 39.53 
                    & 15.39 
                    & 15.95 \\
& w$\backslash$o S  & 98.16  % taum 10 taus 2.5
                    & 86.40
                    & 88.86 
                    & 85.80 
                    & 87.03 
                    & 80.34
                    & 80.54 \\

& \textbf{DGN}  
                & \textbf{98.59}     % Clean
                & \textbf{95.34}     % Addition Noise   0.012
                & \textbf{93.70}     % Subtractive Noise    0.4
                & \textbf{95.43}     % Mixed Noise  0.2
                & \textbf{90.35}     % FGSM Attack  0.006
                & \textbf{86.76}     % PGD Attack   0.003
                & \textbf{86.88} \\  % BIM Attack   0.006
\midrule
\multirow{3}{*}{\rotatebox{90}{Rec}} 
% & LIF               & xx.xx 
%                     & xx.xx 
%                     & xx.xx 
%                     & xx.xx 
%                     & xx.xx 
%                     & xx.xx 
%                     & xx.xx \\
& w$\backslash$o D    & 97.80 
                    & 73.23
                    & 79.22 
                    & 67.88 
                    & 26.65 
                    & 61.79
                    & 60.7 \\
& w$\backslash$o S  & 98.62     % taum 10000 taus 2.5 
                    & 93.47 
                    & 95.61
                    & 91.83 
                    & 82.70
                    & 74.48 
                    & 73.46 \\
                    
& \textbf{DGN}  
                & \textbf{99.10}     % Clean
                & \textbf{94.84}     % Addition Noise   0.012
                & \textbf{94.36}     % Subtractive Noise        0.4
                & \textbf{93.86}     % Mixed Noise
                & \textbf{89.40}     % FGSM Attack  0.006
                & \textbf{87.52}     % PGD Attack   0.003
                & \textbf{87.68} \\  % BIM Attack   0.006
\midrule
\end{tabular}
\label{tab:ablation_study}
\end{table}

To systematically investigate the synergistic interaction between static and dynamic components in the gating architecture, we performed comprehensive ablation studies on the TIDIGITS dataset. As evidenced in Tab. \ref{tab:ablation_study}, three configurations were evaluated: the baseline DGN model, its static-component-deprived variant (w/o S), and dynamic-component-deprived counterpart (w/o D). % Experimental results demonstrate that models incorporating dynamic gating mechanisms exhibit significantly superior classification accuracy and noise robustness compared to those relying solely on static gate structures. This empirical evidence suggests that while static gates provide fundamental information screening capabilities, they lack the adaptive capacity required for dynamic noise suppression. In contrast, dynamic gating architectures enable data-driven learning of context-sensitive "information filters" through parameter adaptation, optimizes propagation of task-relevant features through adaptive feature selection, and actively suppresses noise interference during information transmission. More importantly, the hierarchical integration of static and dynamic gating components establishes a complementary architecture where foundational filtering mechanisms interact synergistically with adaptive modulation pathways. This structural synergy not only preserves essential information screening functions but also introduces context-aware refinement capabilities, ultimately leading to enhanced model performance in complex classification scenarios.
Experimental data show that the model with a dynamic gate structure is significantly higher in classification accuracy and noise resistance than the model with only a static gate structure. This result shows that the static gate structure can provide the model with a basic information screening capability, but cannot dynamically filter noise information. The dynamic gate structure can provide a more free, flexible and adaptable "information selection gate" based on the basic information filtering capability by learning the data information, enhancing the efficiency of effective information transmission while filtering out useless information such as noise. The joint synergy of the static gate structure and the dynamic gate structure further improves the model performance.
% To elucidate the complementary roles of static and dynamic components in gating mechanisms, we conducted ablation studies on the TIDIGITS dataset, comparing the baseline DGN model with its static-free (w/o S) and dynamic-free (w/o D) variants. As shown in Table \ref{tab:ablation_study}, models incorporating dynamic gates achieved superior accuracy and noise robustness. These findings indicate that while static gates offer foundational filtering, they lack adaptivity. In contrast, dynamic gating enables context-sensitive modulation via data-driven parameter tuning. Crucially, integrating both components forms a synergistic architecture that preserves essential filtering while enhancing task-relevant information flow, yielding improved performance in noisy classification tasks.

% 遗憾的是，由于计算资源和时间的限制，我们未能进行更广泛的实验。这是我们当前研究的一个关键限制，因为它阻碍了我们得出具有统计显著性的结论。解决这一限制将是未来工作的一个重要方向。尽管如此，我们的模型在多种类型的噪声和攻击方法下，在多个数据集上表现出了持续的强劲性能，这在一定程度上弥补了这一缺陷。
\subsection{Limitation} \label{limit}
% It is regrettable that, due to constraints in computational resources and time, we were unable to conduct more extensive experiments. This represents a key limitation of our current study, as it prevents us from drawing statistically significant conclusions. Addressing this limitation will be an important direction for future work. Nonetheless, our model exhibits consistently strong performance under various types of noise and attack methods across multiple datasets, which to some extent compensates for this shortcoming.
While the present study contributes valuable insights into the robustness properties of the proposed DGN model, several limitations warrant acknowledgment. First, the investigation primarily focused on robustness analysis, leaving other computational characteristics such as temporal dynamics unexplored due to time constraints. Second, although our neuron design demonstrates marked improvements in biological plausibility and functional performance compared to traditional LIF models, persistent discrepancies remain in replicating key neurobiological features observed in biological neurons. 
% Third, due to time and computational constraints, the practical application of our model in large-scale neural networks and complex real-world scenarios requires further investigation to evaluate its scalability and adaptability. These limitations highlight critical directions for future research to fully characterize the model's capabilities and biological fidelity.

\end{document}